\documentclass{article}

\PassOptionsToPackage{numbers,sort&compress}{natbib}
\PassOptionsToPackage{table}{xcolor}

 \usepackage[main, final]{neurips_2026}
\usepackage[utf8]{inputenc} % allow utf-8 input
\usepackage[T1]{fontenc}    % use 8-bit T1 fonts
\usepackage{hyperref}       % hyperlinks
\usepackage{url}            % simple URL typesetting
\usepackage{booktabs}       % professional-quality tables
\usepackage{amsfonts}       % blackboard math symbols
\usepackage{amsmath}
\usepackage{amssymb}
\usepackage{nicefrac}       % compact symbols for 1/2, etc.
\usepackage{microtype}      % microtypography
\usepackage{xcolor}         % colors
\usepackage{graphicx}
\usepackage{float}
\usepackage{subcaption}
\usepackage{pifont}
\usepackage{algorithm}
\usepackage{algpseudocode}
\usepackage{titletoc}

\DeclareMathOperator*{\argmin}{arg\,min}
\DeclareMathOperator*{\argmax}{arg\,max}

\newcommand{\JI}{\mathrm{JI}_{\text{bona}}}
\newcommand{\JER}{\mathrm{JER}_{\text{spoof}}}
\newcommand{\Lassign}{\mathcal{L}_{\text{assign}}}
\newcommand{\Lexist}{\mathcal{L}_{\text{exist}}}
\newcommand{\Ltotal}{\mathcal{L}_{\text{total}}}
\newcommand{\Lbin}{\mathcal{L}_{\text{bin}}}
\newcommand{\Kmax}{K_{\max}}

\usepackage{booktabs}
\usepackage{threeparttable}
\usepackage{multirow}
\usepackage{makecell}
\usepackage{colortbl}
\usepackage{array}

\definecolor{oursbg}{HTML}{E8F0FB}   % row of the proposed model
\definecolor{groupbg}{HTML}{F3F4F6}  % group header rows
\definecolor{gaingreen}{HTML}{15803D}
\definecolor{lossred}{HTML}{B91C1C}
\definecolor{mutedgray}{HTML}{6B7280}
\definecolor{known}{HTML}{2563EB}
\definecolor{varied}{HTML}{0D9488}
\definecolor{unknown}{HTML}{DC2626}

\newcommand{\ms}[2]{#1{\scriptsize\,$\pm$#2}}
\newcommand{\best}[1]{\textbf{#1}}
\newcommand{\second}[1]{\underline{#1}}
\newcommand{\gain}[1]{{\scriptsize\color{gaingreen}#1}}
\newcommand{\loss}[1]{{\scriptsize\color{lossred}#1}}
\newcommand{\oursrow}{\rowcolor{oursbg}}
\newcommand{\grouprow}[2]{\rowcolor{groupbg}\multicolumn{#1}{l}{\textit{\small #2}}\\}
\newcommand{\nr}{{\color{mutedgray}n/r}}
\newcommand{\na}{{\color{mutedgray}--}}
\newcommand{\Kn}[1]{{\color{known}#1}}
\newcommand{\Va}[1]{{\color{varied}#1}}
\newcommand{\Un}[1]{{\color{unknown}#1}}
\newcommand{\tcite}[1]{{\scriptsize\citep{#1}}}
\newcommand{\tabnote}[1]{\par\vspace{3pt}{\footnotesize\noindent #1\par}}

\titlecontents{psection}[1.6em]
  {\addvspace{0.55em}\bfseries}
  {\contentslabel{1.6em}}
  {\hspace*{-1.6em}}
  {\hfill\contentspage}
\titlecontents{psubsection}[4.2em]
  {\addvspace{0.05em}}
  {\contentslabel{2.4em}}
  {\hspace*{-2.4em}}
  {\hfill\contentspage}

\definecolor{linkblue}{HTML}{1F4E9A}
\hypersetup{colorlinks=true, linkcolor=linkblue, citecolor=linkblue, urlcolor=linkblue}

\title{Clustering-Free End-to-End Spoof Diarization with Encoder-Decoder Attractors}

\author{%
  Bongsu Jung\textsuperscript{1} \quad Donghee Kim\textsuperscript{2} \quad Wooil Kim\textsuperscript{1,2} \\
  \textsuperscript{1}Department of Computer Science and Engineering \\
  \textsuperscript{2}Department of Artificial Intelligence \\
  Incheon National University \\
  \texttt{\{qhd0330, d0\_0h, wikim\}@inu.ac.kr}
}

\begin{document}

\maketitle

\begin{abstract}
Spoof diarization aims to jointly localize spoofed regions and determine which spoofing method generated each segment within a partially spoofed utterance.
Existing approaches are modular pipelines that separate localization and clustering.
This separation causes a training--inference mismatch, and the clustering uses the true number of spoofing classes at inference.
We propose E2E-SD, to our knowledge the first clustering-free end-to-end framework for spoof diarization, which reformulates the task as variable-cardinality temporal set prediction.
E2E-SD jointly learns localization and class assignment using encoder-decoder attractors with Hungarian-matched training, eliminating the need for post-hoc clustering and oracle class counts.
Our framework dynamically generates class-specific attractors through an LSTM decoder and estimates the number of active spoofing classes via attractor existence probabilities.
On the PartialSpoof benchmark, E2E-SD achieves a spoof Jaccard error rate ($\JER$) of 16.01\,$\pm$\,0.95\% over three runs without oracle class counts.
This is 34.3\% lower than the previous best result (24.38\%), and the number of classes is estimated with 88.4\% accuracy.
The largest gain is observed on unknown attacks, where $\JER$ drops from 36.78\% to 22.33\% (39.3\% relative improvement), suggesting that end-to-end optimization with dynamic attractors enables more attack-agnostic generalization than modular clustering pipelines.
Code available at: \mbox{\url{https://github.com/Bongsujung/E2E-SD}}
\end{abstract}

% ============================================================
\section{Introduction}
\label{sec:intro}

% --- Paragraph 1: Threat landscape & task hierarchy ---
Recent advances in neural speech synthesis and voice conversion, exemplified by
zero-shot voice cloning from only a few seconds of enrolled
speech~\citep{wang2023valle}, have lowered the barrier to producing
convincing manipulations~\citep{yi2023addsurvey}. Adversaries need not synthesize
an entire utterance; manipulating only a few words or short segments suffices to alter
meaning while evading utterance-level detectors~\citep{zhang2023partialspoof,yi2021halfttruth}.
Current detection systems exhibit severe generalization failures on real-world deepfakes~\citep{mueller2022generalize}
and may rely on synthesis artifacts rather than fundamental spoofing
characteristics~\citep{shih2024artifacts}.
Partially spoofed audio, containing both bonafide and synthetic segments, has its own
tracks in the ADD 2022 and 2023 challenges~\citep{yi2022add,yi2023add2}.

Defending against such attacks requires progressively deeper analysis.
Spoof detection performs utterance-level binary
classification~\citep{jung2022aasist,tak2021rawnet2,tak2022automatic};
spoof localization extends this to the frame level to identify \textit{when}
spoofing occurs~\citep{zhang2021initial,wu2022fakesspan,cai2023boundary,zhang2023rangeeer,zhong2024boundary,mao2026sal,kim2026transition}.
Yet neither task reveals \textit{what} spoofing method was used, information essential
for forensic source attribution~\citep{muller2022attacker,klein2024source,klein2025openset}.
Spoof diarization, formalized by~\citet{zhang2024spoofdia}, answers ``what spoofed when''
by jointly localizing spoofed segments and clustering them by generating method.
Analogous to speaker diarization~\citep{park2022review} (``who spoke when''),
it differs in two ways: spoofed segments do not overlap, and the system
must handle an unknown number of unseen spoofing methods (open-set challenge).

% --- Paragraph 2: Existing approaches and limitations ---
The 3C benchmark model~\citep{zhang2024spoofdia} decomposes the task into a
localization branch and a diarization branch with agglomerative hierarchical
clustering~(AHC), fused via a Label-based CM-constraint (LCM).
\citet{koo2025token} improved clustering via learnable attractor tokens with
cross-attention and a merged single-branch architecture, and trained their models with
P2SGrad~\citep{zhang2019p2sgrad}.
The two approaches share three structural limitations.
(1)~Oracle $K$ dependency: both models run AHC with the true number of spoofing
classes at inference, which is unknown in practice.
(2)~Training--inference mismatch: the embeddings are trained with
frame-level classification losses but evaluated via post-hoc clustering.
(3)~Static class representation: neither model generates one class representation
per spoofing method in an utterance; \citet{koo2025token} use the same two attractor
tokens (bonafide and spoof) for every utterance.

% --- Paragraph 3: Our method and contributions ---
We observe that spoof diarization is a variable-cardinality single-label
assignment problem, admitting an end-to-end formulation analogous to set prediction
in object detection~\citep{carion2020detr} and the EEND-EDA framework in speaker
diarization~\citep{horiguchi2020endtoend,horiguchi2022encoder}.
We propose \textbf{E2E-SD}, to our knowledge the first clustering-free end-to-end framework for spoof diarization.
It reformulates the task within the encoder-decoder attractor (EDA) framework, which derives an utterance-specific attractor for each class.
Three modifications adapt the EDA to the single-label spoof setting: (i)~softmax cross-entropy over a cosine similarity matrix, whose argmax is also the inference rule; (ii)~learnable
affine scaling of cosine similarity inspired by GE2E~\citep{wan2018ge2e} to improve class discrimination; and (iii)~joint
class-count estimation via attractor existence probabilities, which removes the oracle $K$.
Our contributions are:
\begin{itemize}
    \item We formulate spoof diarization as variable-cardinality temporal set
    prediction and propose E2E-SD, replacing clustering with Hungarian-matched
    assignment training and argmax inference.
    \item We demonstrate oracle-free class-count estimation via EDA existence
    probabilities, achieving 88.4\% accuracy and a spoof Jaccard error rate ($\JER$)
    within 0.1 points of oracle~$K$.
    \item On the PartialSpoof benchmark~\citep{zhang2023partialspoof}, E2E-SD achieves
    $\JER$~=~16.01\,$\pm$\,0.95\% over three runs, a 34.3\% relative improvement over the best prior result
    (24.38\%, with oracle $K$~\citep{koo2025token}).
\end{itemize}

% ============================================================
\section{Related work}
\label{sec:related}

\paragraph{Partially spoofed audio detection and localization.}
Earlier ASVspoof challenges focused on fully synthesized or replayed speech~\citep{wu2015asvspoof,kinnunen2017asvspoof,yamagishi2021asvspoof}; partial spoofing was first studied systematically by~\citet{zhang2021initial} and~\citet{yi2021halfttruth}. The PartialSpoof dataset~\citep{zhang2021initial,zhang2023partialspoof}, constructed from ASVspoof 2019 LA~\citep{yamagishi2019asvspoof,todisco2019asvspoof}, provides segment-level ground truth at resolutions down to 20\,ms and is the standard benchmark for this setting. Countermeasures (CMs) trained on fully spoofed data degrade under partial spoofing~\citep{zhang2021initial,zhang2023partialspoof}, motivating frame-level approaches including fake span discovery~\citep{wu2022fakesspan}, boundary detection~\citep{cai2023boundary,zhong2024boundary}, coarse-to-fine proposal refinement~\citep{wu2024cfprf}, segment-aware learning~\citep{mao2026sal}, and transition-aware modeling~\citep{kim2026transition}. CM analysis via Grad-CAM revealed that models trained on PartialSpoof focus on the artifacts at the transitions between segments~\citep{liu2024spoofcm}. The ADD 2022 and 2023 challenges~\citep{yi2022add,yi2023add2} added partial-spoof tracks, and LlamaPartialSpoof~\citep{luong2025llamapartialspoof} and PartialEdit~\citep{zhang2025partialedit} evaluate localization outside PartialSpoof. We extend frame-level analysis from binary localization to multi-class diarization.
\paragraph{Spoof diarization.}
\citet{zhang2024spoofdia} formalized spoof diarization and its metrics, $\JI$ and $\JER$. Their 3C benchmark consists of a diarization branch (CM-dia embedding extractor $+$ AHC) and a localization branch (CM-loc, a binary or multi-class countermeasure), fused via the LCM. Both branches share the same wav2vec2-Large~\citep{baevski2020wav2vec} $+$ gMLP architecture but are trained separately under different labeling schemes and combined at inference, requiring two full model instances. \citet{koo2025token} improved upon 3C by introducing two learnable attractor tokens that interact with frame embeddings via cross-attention, and proposed a merged single-branch architecture using a single XLS-R encoder. All their models are trained with P2SGrad loss~\citep{zhang2019p2sgrad}, and the tokens yield relative $\JER$ improvements of up to 17.2\%. However, both works retain AHC with oracle $K$ and do not report results in the oracle-free setting.

\paragraph{End-to-end neural diarization.}
End-to-end neural diarization (EEND)~\citep{fujita2019end} reformulates speaker diarization as frame-level multi-label classification with permutation invariant training (PIT)~\citep{yu2017pit}, enabling joint optimization of speech activity detection and speaker assignment. This eliminates the traditional pipeline of i-vector or x-vector embedding extraction~\citep{dehak2011ivector,snyder2018xvector} followed by AHC or spectral clustering. Self-attentive EEND~\citep{fujita2019saend} replaces the original bidirectional LSTM with a Transformer and improves accuracy. EEND-EDA~\citep{horiguchi2020endtoend,horiguchi2022encoder} extends this with an LSTM encoder-decoder attractor module that dynamically generates speaker-specific attractors and estimates the number of speakers via existence probabilities. Subsequent work introduced Transformer-based attractor decoders~\citep{samarakoon2023transformer}, cross-attention between attractors and frame embeddings for enhanced discrimination~\citep{chen2023aed}, Perceiver-based attractors~\citep{landini2024diaper}, global-local attractor structures for handling unlimited numbers of speakers~\citep{horiguchi2021global}, and hybrid end-to-end plus clustering approaches~\citep{kinoshita2021integrating}. E2E-SD adopts the EDA framework. Because spoofed segments do not overlap, it replaces per-speaker binary cross-entropy (BCE) with multi-class cross-entropy and PIT with Hungarian matching~\citep{kuhn1955hungarian}.

% ============================================================
\section{Method}
\label{sec:method}

\subsection{Problem formulation}
\label{sec:formulation}

Given a waveform $\mathbf{x} \in \mathbb{R}^{T_{\text{raw}}}$, spoof diarization assigns each frame $t \in \{1, \ldots, T\}$ a label $c_t \in \{0, 1, \ldots, K{-}1\}$, where class~$0$ denotes bonafide and classes $1, \ldots, K{-}1$ correspond to distinct spoofing methods. The number of classes $K$ varies per utterance, from $K{=}1$ (fully bonafide) up to $K{=}11$ in the PartialSpoof evaluation set: bonafide, up to 9 co-occurring spoofing methods from the 13 ASVspoof 2019 LA evaluation attacks, and the concatenated parts (ConP), which PartialSpoof labels as a separate class. We cast this as variable-cardinality temporal set prediction: the system must jointly estimate the class count~$\hat{K}$ and produce a label sequence $\{\hat{c}_t\}_{t=1}^{T}$, without knowing~$K$ in advance. Three structural properties shape our design:
\begin{enumerate}
    \item \textbf{Single-label assignment:} each frame belongs to exactly one class (spoofed segments do not overlap), whereas in speaker diarization multiple speakers may be active simultaneously.
    \item \textbf{Open-set classes:} spoofing methods in the test set may be entirely unseen during training, demanding representations that generalize beyond known attacks. E2E-SD has no output layer tied to the training attacks: its attractors are generated from each utterance.
    \item \textbf{Asymmetric class structure:} the label space exhibits an inherent two-level grouping: bonafide vs.\ spoof at the top level, and which spoofing method at the bottom. As noted in~\citet{zhang2024spoofdia}, spoof diarization considers ``two primary groups of clusters: bona fide and spoof.'' Bonafide is present in almost all utterances (only 2,153 of the 71,237 evaluation utterances contain no bonafide segments), and its semantics are consistent regardless of which spoofing methods co-occur. This asymmetry motivates fixing the first attractor to bonafide as a structural prior (Section~\ref{sec:assignment}).
\end{enumerate}

\subsection{Architecture overview}
\label{sec:overview}

\begin{figure}[t]
  \centering
  \includegraphics[width=\linewidth]{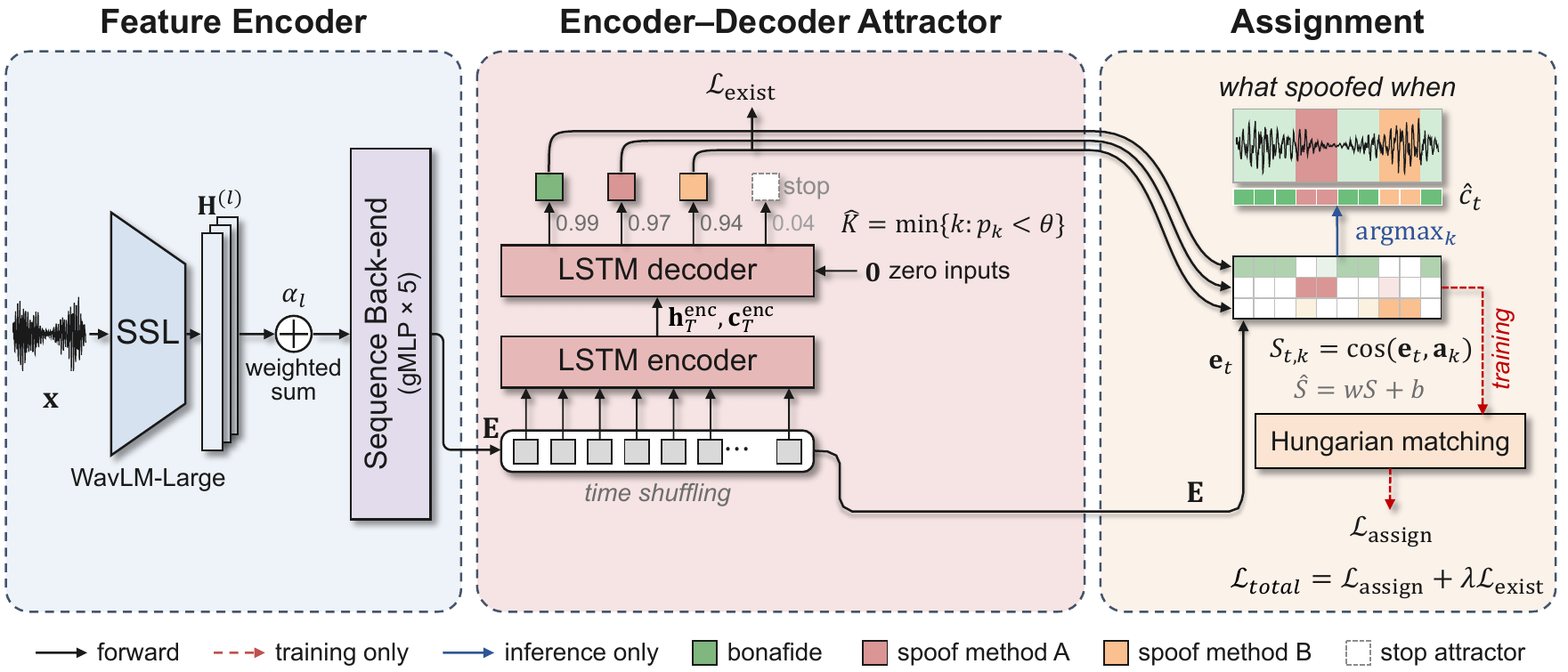}
  \caption{Overview of the E2E-SD architecture. The feature encoder produces frame embeddings $\mathbf{E}$. The EDA head generates class-specific attractors $\mathbf{A}$ and existence probabilities $\mathbf{p}$, with attractor $\mathbf{a}_0$ fixed to bonafide (green). Cosine similarity between $\mathbf{E}$ and $\mathbf{A}$ forms the assignment matrix $\mathbf{S}$, trained with Hungarian-matched cross-entropy. At inference, $\hat{K}$ is estimated from $\mathbf{p}$ and frames are assigned via argmax.}
  \label{fig:architecture}
\end{figure}

E2E-SD consists of three components (Figure~\ref{fig:architecture}):
(1)~a feature encoder (Section~\ref{sec:encoder}) that extracts frame-level embeddings $\mathbf{E} \in \mathbb{R}^{T \times d}$ from the raw waveform;
(2)~an encoder-decoder attractor (EDA) head (Section~\ref{sec:eda}) that generates class-specific attractors $\mathbf{A} \in \mathbb{R}^{K_{\max} \times d}$ and existence probabilities $\mathbf{p} \in [0,1]^{K_{\max}}$; and
(3)~an assignment module (Section~\ref{sec:assignment}) that computes cosine similarity between frames and attractors and trains via Hungarian-matched cross-entropy.

\subsection{Feature encoder}
\label{sec:encoder}

\paragraph{SSL front-end.}
We use a pre-trained self-supervised learning (SSL) model (WavLM-Large~\citep{chen2022wavlm} by default) that produces $L$ hidden-state sequences $\{\mathbf{H}^{(l)} \in \mathbb{R}^{T \times d_{\text{ssl}}}\}_{l=0}^{L-1}$ at 20\,ms frame resolution ($d_{\text{ssl}} = 1024$). Since SSL layers differ in how useful they are for deepfake detection~\citep{elkheir2025comprehensive}, we compute a learnable weighted sum:
\begin{equation}
    \mathbf{H} = \sum_{l=0}^{L-1} \alpha_l \mathbf{H}^{(l)}, \quad \alpha_l = \frac{\exp(w_l)}{\sum_{l'=0}^{L-1} \exp(w_{l'})},
    \label{eq:weighted_sum}
\end{equation}
where $\{w_l\}$ are learnable scalars. The SSL encoder is fine-tuned end-to-end with the same base learning rate as the rest of the network.

\paragraph{Sequence back-end.}
The weighted features are processed by a 5-layer sequence model to produce task-specific frame embeddings $\mathbf{E} = f_{\text{backend}}(\mathbf{H}) \in \mathbb{R}^{T \times d}$, where $d = 512$. We adopt gMLP~\citep{liu2021gmlp} as the default back-end, consistent with prior PartialSpoof work~\citep{zhang2024spoofdia,koo2025token}, and evaluate Transformer and Conformer alternatives in the ablation study.

\subsection{Encoder-decoder attractor}
\label{sec:eda}

We reformulate spoof diarization within the LSTM-based encoder-decoder attractor (EDA) framework~\citep{horiguchi2020endtoend}. The EDA takes the frame embeddings $\mathbf{E}$ and produces class-specific attractors: vectors that act as prototypes in the same embedding space, each representing a distinct class.

\paragraph{Attractor generation.}
The attractor encoder is an LSTM that summarizes the full frame sequence into a final hidden state. Following EEND-EDA~\citep{horiguchi2022encoder}, we apply time shuffling during training and inference: the temporal order of frames is randomly permuted before encoding. This prevents the LSTM from exploiting temporal position and encourages it to encode class-identity information from frame content:
\begin{equation}
    \mathbf{h}_t^{\text{enc}},\, \mathbf{c}_t^{\text{enc}} = \text{LSTM}_{\text{enc}}(\mathbf{e}_{\psi_t},\, \mathbf{h}_{t-1}^{\text{enc}},\, \mathbf{c}_{t-1}^{\text{enc}}), \quad t = 1, \ldots, T,
    \label{eq:eda_enc}
\end{equation}
where $(\psi_1, \ldots, \psi_T)$ is a random permutation of $(1, \ldots, T)$, drawn anew at each training step and generated for each utterance with a fixed random seed at inference. The encoder states are zero-initialized: $\mathbf{h}_0^{\text{enc}} = \mathbf{c}_0^{\text{enc}} = \mathbf{0}$. The LSTM decoder is initialized with the encoder's final hidden and cell states and generates attractors from zero-input tokens:
\begin{equation}
    \mathbf{a}_k,\, (\mathbf{h}_k^{\text{dec}},\, \mathbf{c}_k^{\text{dec}}) = \text{LSTM}_{\text{dec}}(\mathbf{0},\, \mathbf{h}_{k-1}^{\text{dec}},\, \mathbf{c}_{k-1}^{\text{dec}}), \quad k = 0, 1, \ldots.
    \label{eq:eda_dec}
\end{equation}
At inference, the decoder generates $K_{\max}$ attractors for every utterance. During training, for an utterance with $K$ classes, the $(K{+}1)$-th attractor acts as a \textit{stop attractor} whose existence target is $0$ (App.~\ref{app:config}).

\paragraph{Existence estimation.}
A linear existence head predicts whether each attractor corresponds to a real class:
\begin{equation}
    p_k = \sigma(\mathbf{v}^\top \mathbf{a}_k + b_{\text{cnt}}), \qquad
    \Lexist = -\frac{1}{K{+}1} \sum_{k=0}^{K} \big[ y_k \log p_k + (1 - y_k) \log(1 - p_k) \big],
    \label{eq:exist}
\end{equation}
where $y_k = \mathbf{1}[k < K]$ is the existence target, $\mathbf{1}[\cdot]$ is the indicator function, and $\sigma$ is the sigmoid. The existence probabilities enable oracle-free class-count estimation at inference (Section~\ref{sec:inference}).

\subsection{Assignment via Hungarian matching and scaled cosine similarity}
\label{sec:assignment}

\paragraph{Frame-attractor similarity.}
Frame embeddings and attractors are $\ell_2$-normalized; their cosine similarity forms the assignment matrix $\mathbf{S} \in \mathbb{R}^{T \times K}$:
\begin{equation}
    S_{t,k} = \frac{\mathbf{e}_t^\top \mathbf{a}_k}{\|\mathbf{e}_t\| \cdot \|\mathbf{a}_k\|} \in [-1, 1].
    \label{eq:cos_sim}
\end{equation}

\paragraph{Hungarian matching with bonafide anchoring.}
Training requires an optimal bipartite matching between attractors and ground-truth classes~\citep{carion2020detr,yu2017pit}. We fix $\mathbf{a}_0$ to bonafide and apply the Hungarian algorithm over the remaining $K{-}1$ spoof classes only. This anchoring exploits the asymmetric class structure of spoof diarization (Section~\ref{sec:formulation}) and reduces the search space from $K!$ to $(K{-}1)!$. The cost matrix $\mathbf{C} \in \mathbb{R}^{(K-1) \times (K-1)}$ has entries $C_{n,k} = -\sum_{t:\, c_t = n} S_{t,k}$ for $n, k \in \{1, \ldots, K{-}1\}$; the optimal permutation is:
\begin{equation}
    \pi^* = \argmin_{\pi \in \Pi_{K-1}} \sum_{n=1}^{K-1} C_{n, \pi(n)},
    \quad \pi^*(0) = 0,
    \label{eq:hungarian}
\end{equation}
yielding a per-frame target $y_t = \pi^*(c_t)$. For the few utterances without bonafide speech, the matching runs over all classes. Because every attractor is generated per utterance from its own frame embeddings (Eqs.~\ref{eq:eda_enc}--\ref{eq:eda_dec}), $\mathbf{a}_0$ adapts to the speaker and channel of the utterance; the anchoring only reserves index~0 during matching.

\paragraph{Scaled cosine similarity.}
Since $S_{t,k} \in [-1,1]$, direct softmax produces near-uniform probability distributions, limiting class discrimination. Following GE2E~\citep{wan2018ge2e}, which uses a learnable affine transformation of the cosine similarity, we apply:
\begin{equation}
    \hat{S}_{t,k} = wS_{t,k} + b,
    \label{eq:scaled_sim}
\end{equation}
where $w$ and $b$ are learnable scalars, initialized at $w{=}10$ and $b{=}{-}5$.

The assignment loss over valid frames $\mathcal{V} = \{t : c_t \geq 0\}$ is:
\begin{equation}
    \Lassign = -\frac{1}{|\mathcal{V}|}\sum_{t \in \mathcal{V}} \log \frac{\exp(\hat{S}_{t,\,\pi^*(c_t)})}{\sum_{k=0}^{K-1} \exp(\hat{S}_{t,k})},
    \label{eq:assign}
\end{equation}
where non-speech frames other than ConP ($c_t{=}{-}1$) are excluded from the loss, following 3C, and the remaining frames carry $c_t \in \{0,\ldots,K{-}1\}$.

\subsection{Design differences from speaker diarization}
\label{sec:domain}

Table~\ref{tab:adaptation} summarizes the changes needed to apply the EDA framework to spoof diarization.
\begin{table}[t]
\centering\small
\caption{Differences in applying EDA to speaker vs.\ spoof diarization.}
\label{tab:adaptation}
\begin{tabular}{@{}lll@{}}
\toprule
Component & Speaker diarization (EEND-EDA) & Spoof diarization (E2E-SD) \\
\midrule
Overlap & Multi-label (speakers overlap) & Single-label (classes exclusive) \\
Frame loss & Per-class sigmoid + BCE & Softmax CE (Eq.~\ref{eq:assign}) \\
Similarity & Dot product (unbounded) & Cosine + affine (Eq.~\ref{eq:scaled_sim}) \\
Matching & PIT over all $K!$ permutations & Hungarian, $\mathbf{a}_0$ fixed to bonafide \\
\bottomrule
\end{tabular}
\end{table}

The single-label structure enables competition among classes via softmax: each frame's gradient simultaneously reinforces the correct attractor and suppresses all others. Hungarian matching and argmax inference therefore require no thresholding or post-processing for the frame labels.

\subsection{Training objective and inference}
\label{sec:inference}

\paragraph{Training.} The total objective combines the assignment and existence losses:
\begin{equation}
    \Ltotal = \Lassign + \lambda \Lexist, \quad \lambda = 1.0.
    \label{eq:total}
\end{equation}

\paragraph{Inference.}
We support two inference modes. In oracle mode, the ground-truth $K$ is provided and exactly $K$ attractors are used. In estimated-$K$ mode (the practical setting), the class count is estimated from existence probabilities:
\begin{equation}
    \hat{K} = \min\{k : p_k < \theta\}, \quad \theta = 0.5,
    \label{eq:auto_k}
\end{equation}
defaulting to $\hat{K} = K_{\max}$ if all probabilities exceed $\theta$. Frame assignments are then obtained by argmax over the scaled similarity: $\hat{c}_t = \argmax_{k \in \{0, \ldots, \hat{K}-1\}} \hat{S}_{t,k}$. This procedure requires no clustering, no post-processing, and no oracle information.

% ============================================================
\section{Experiments}
\label{sec:experiments}

\subsection{Dataset and evaluation protocol}
\label{sec:protocol}

We train and evaluate on the PartialSpoof dataset~\citep{zhang2023partialspoof}, the only publicly available benchmark with segment-level multi-class annotations for spoof diarization. The dataset contains 25,380 / 24,844 / 71,237 utterances for train / dev / eval, constructed by concatenating bonafide and spoofed segments of ASVspoof 2019 LA~\citep{yamagishi2019asvspoof} (attacks A01--A06 in train and dev, and the 13 attacks A07--A19 in eval). Training uses only the PartialSpoof training set, without any other corpus or data augmentation. Following~\citet{zhang2024spoofdia}, attacks are grouped by their relation to the training attacks: Known (A16, A19, identical to a training attack), Varied (A07--A09, A17, sharing the acoustic model or waveform generator with a training attack), and Unknown (A10--A15, A18, sharing neither). We report $\JI$ (bonafide localization error, averaged over the utterances that contain bonafide speech), $\JER$ (spoof Jaccard error rate), per-group $\JER$, and class-count estimation accuracy ($K$-acc.; App.~\ref{app:metrics}).\footnote{We compute the metrics of~\citet{zhang2024spoofdia} with dscore~\citep{dscore} at a 0.1\,ms step over the reference speech.} All baseline numbers are taken directly from the original publications~\citep{zhang2024spoofdia,koo2025token}; we do not re-implement baseline models (App.~\ref{app:protocol} compares the protocols).

\subsection{Implementation details}
\label{sec:impl}

The default configuration uses WavLM-Large~\citep{chen2022wavlm} with a 5-layer gMLP back-end ($d{=}512$) and 1-layer LSTM encoder/decoder (hidden size 512); $K_{\max}{=}8$ is the largest number of co-occurring classes in the PartialSpoof training and development labels, and for another training corpus $K_{\max}$ is set by the same rule. Models are trained with Adam~\citep{kingma2015adam} (base lr~$=10^{-5}$, including the SSL model) and early stopping on the dev-set loss; App.~\ref{app:config} lists the full configuration. Training takes about 6~hours on a single NVIDIA RTX 4090 GPU. For every configuration, we report mean\,$\pm$\,std over three runs.

\subsection{Main results}
\label{sec:main_results}

\begin{table}[t]
  \caption{Main results on the PartialSpoof eval set (\%). Prior methods use oracle $K$; E2E-SD operates fully oracle-free (estimated $K$) and is reported as mean\,$\pm$\,std over three runs. The E2E-SD rows with w2v2-L and XLS-R use the SSL front-ends of the baselines. \best{Bold}: best; \second{underline}: second best. Per-group $\JER$ for \citet{koo2025token} is not reported (n/r).}
  \label{tab:main}
  \centering
  \small
  \setlength{\tabcolsep}{3.2pt}
  \renewcommand{\arraystretch}{1.12}
  % mean right-aligned in a fixed box, std in a fixed box: columns align, no overlap with \rowcolor
  \newcommand{\mm}[2]{\makebox[2.3em][r]{#1}\makebox[2.05em][l]{\,{\scriptsize$\pm$#2}}}
  \newcommand{\mo}[1]{\makebox[2.3em][r]{#1}\makebox[2.05em][l]{}}
  \begin{threeparttable}
  \begin{tabular}{llcccccc}
    \toprule
    & & & & \multicolumn{3}{c}{$\JER$ by attack group $\downarrow$} & \\
    \cmidrule(lr){5-7}
    Method & SSL & $\JI\downarrow$ & $\JER\downarrow$ & \Kn{Known} & \Va{Varied} & \Un{Unknown} & $K$-acc.\,$\uparrow$ \\
    \midrule
    \grouprow{8}{3C baseline~\citep{zhang2024spoofdia}: AHC, oracle $K$}
    Mul-CM / --      & w2v2-L & \mo{19.66} & \mo{28.05} & \mo{11.56} & \mo{15.49} & \mo{36.78} & \na \\
    Mul-CM / Bin-CM  & w2v2-L & \mo{15.15} & \mo{34.13} & \mo{11.98} & \mo{15.49} & \mo{49.02} & \na \\
    Mul-CM / Mul-CM  & w2v2-L & \mo{17.08} & \mo{35.34} & \mo{11.96} & \mo{16.25} & \mo{50.64} & \na \\
    \grouprow{8}{Attractor token~\citep{koo2025token}: AHC, oracle $K$}
    3C$^*$ + Attractor & XLS-R & \mo{15.58} & \mo{24.38} & \nr & \nr & \nr & \na \\
    Merged + Attractor & XLS-R & \mo{18.05} & \mo{26.99} & \nr & \nr & \nr & \na \\
    \grouprow{8}{E2E-SD (ours): no clustering, estimated $K$}
    E2E-SD & w2v2-L & \mm{11.56}{0.39} & \mm{21.90}{0.56} & \mm{12.26}{0.68} & \mm{12.77}{0.66} & \mm{30.25}{0.47} & \mm{\second{84.53}}{1.20} \\
    E2E-SD & XLS-R & \mm{\second{10.77}}{1.74} & \mm{\second{19.60}}{1.60} & \mm{\best{7.94}}{0.19} & \mm{\second{10.18}}{0.70} & \mm{\second{28.69}}{2.75} & \mm{82.27}{2.44} \\
    \oursrow
    E2E-SD & WavLM-L & \mm{\best{7.77}}{0.68} & \mm{\best{16.01}}{0.95} & \mm{\second{9.55}}{0.42} & \mm{\best{8.70}}{0.24} & \mm{\best{22.33}}{1.77} & \mm{\best{88.44}}{1.65} \\
    \bottomrule
  \end{tabular}
  \tabnote{3C rows: CM-dia / CM-loc (Mul-CM / Bin-CM: best 3C model per \citet{zhang2024spoofdia}); $\JI$, $\JER$: its Table~1; groups: App.\ Table~2 of its arXiv version. 3C$^*$: re-implementation by \citet{koo2025token}. E2E-SD: gMLP back-end, scaled CE. w2v2-L: wav2vec~2.0 Large (3C checkpoint).}
  \end{threeparttable}
\end{table}

Table~\ref{tab:main} presents the main results. Our default configuration (WavLM-L, gMLP, scaled CE) achieves the lowest $\JER$, 16.01\%, against 24.38\% for the best previous result~\citep{koo2025token} (oracle $K$).

\paragraph{Controlled SSL comparison.} Our best results use WavLM-Large, whereas the baselines use wav2vec2-based front-ends. We therefore include a controlled comparison with the SSL models of the baselines: the wav2vec2-Large checkpoint of 3C, and XLS-R 300M for the XLS-R family of~\citet{koo2025token}, whose checkpoint is not stated. With wav2vec2-Large, E2E-SD achieves $\JER$ 21.90\% (estimated $K$) against 28.05\% for the 3C model with the lowest $\JER$ (oracle $K$), 6.15~points (21.9\% relative) lower; with XLS-R, it achieves 19.60\% against 24.38\% for the best attractor-token model, 4.78~points (19.6\% relative) lower. These results suggest that the improvement is primarily attributable to the end-to-end framework.

\paragraph{Unknown attack generalization.} The improvement is most pronounced on unknown attacks: $\JER$ drops from 36.78\% (3C, Mul-CM / --) to 22.33\% with WavLM, 28.69\% with XLS-R and 30.25\% with wav2vec2-Large. The consistent gains across all three SSL front-ends suggest that end-to-end training with dynamic attractors captures more attack-agnostic patterns than modular clustering pipelines.

\subsection{Ablation studies}
\label{sec:ablation}

\paragraph{Assignment loss.} We ablate the main design choices (tables and further ablations in App.~\ref{app:ablations}). Among four losses (CE on cosine, scaled CE ($\hat{S}{=}wS{+}b$), contrastive (GE2E-style), and P2SGrad), scaled CE achieves the best $\JER$ (16.01\%), improving over CE on cosine (18.74\%) by 2.73 points. Learnable affine scaling~\citep{wan2018ge2e} resolves the limited class discrimination of narrow-range cosine logits. P2SGrad~\citep{zhang2019p2sgrad}, the loss used by~\citet{koo2025token}, gives the highest $\JER$ (21.80\%).

\paragraph{Back-end and front-end.} gMLP~\citep{liu2021gmlp} and Transformer~\citep{vaswani2017attention} back-ends yield comparable overall $\JER$ (16.01 vs.\ 17.41\%), while Conformer~\citep{gulati2020conformer} lags behind (18.06\%). WavLM-Large improves over XLS-R (19.60\%) by 3.59 points and over wav2vec2-Large (21.90\%) by 5.89 points, consistent with WavLM's denoising pre-training objective~\citep{chen2022wavlm}.

\paragraph{Oracle vs.\ estimated $K$.} The gap between oracle and estimated $K$ is 0.08 points in $\JER$, validating existence-probability-based class-count estimation. Adding a binary localization head does not improve $\JER$ (17.99 vs.\ 16.01\%) or the 20\,ms segment EER (seg-EER, 15.98 vs.\ 16.02\%; App.~\ref{app:abl_loc}).

\paragraph{Bonafide anchoring.} Removing the bonafide anchor (free Hungarian matching over all $K$ attractors) increases $\JER$ from 16.01\% to 18.50\% (+2.48 points), with the largest degradation on unknown attacks (22.33\% $\to$ 26.51\%). These results suggest that fixing $\mathbf{a}_0$ to bonafide provides a structural prior that stabilizes attractor assignment.

\paragraph{$K_{\max}$ and $\theta$ sensitivity.} $K_{\max}$ and $\theta$ act only at inference (App.~\ref{app:abl_theta}). Any $K_{\max} \in \{6,7,8,11\}$ yields the same results within 0.001 points, and for $\theta \in [0.3, 0.7]$, $\JER$ changes by at most 0.04 points.

\subsection{Transfer beyond PartialSpoof}
\label{sec:transfer}

\begin{table}[t]
  \caption{Transfer beyond PartialSpoof (EER, \%). All models are trained on PartialSpoof only; models trained for localization are a reference for the error level. seg: 20\,ms segment-level EER; utt: utterance EER. E2E-SD: $\mathbf{a}_0$ score, mean over three runs (std below).}
  \label{tab:transfer}
  \centering
  \small
  \setlength{\tabcolsep}{5pt}
  \newcommand{\sd}[1]{{\color{mutedgray}\footnotesize #1}}
  \begin{threeparttable}
  \begin{tabular}{lcccccccc}
    \toprule
    & \multicolumn{2}{c}{Localization (seg) $\downarrow$} & \multicolumn{6}{c}{Detection (utt) $\downarrow$} \\
    \cmidrule(lr){2-3}\cmidrule(l){4-9}
    Model & PS & LlamaPS & PS & LlamaPS & 19LA & 21LA & 21DF & ITW \\
    \midrule
    \grouprow{9}{Trained for localization (reference)}
    MRM~\tcite{zhang2023partialspoof} & 13.70\textsuperscript{a} & 47.49\textsuperscript{a} & 1.48\textsuperscript{a} & 24.49\textsuperscript{a} & \na & \na & \na & \na \\
    CFPRF~\tcite{wu2024cfprf} & 7.41 & 43.25\textsuperscript{b} & 1.72\textsuperscript{b} & 31.50\textsuperscript{b} & \na & \na & \na & \na \\
    reCFPRF~\tcite{luong2025robust} & 9.84 & 41.72 & 1.65 & 24.81 & \na & \na & \na & \na \\
    BAM~\tcite{zhong2024boundary} & 5.20 & 42.58\textsuperscript{c} & \na & \na & \na & \na & \na & \na \\
    SAL (XLS-R)~\tcite{mao2026sal} & \na & 35.52 & \na & \na & \na & \na & \na & \na \\
    \grouprow{9}{Trained for spoof diarization: attractor token~\tcite{koo2025token} and E2E-SD (ours)}
    3C$^*$ + Attractor & 21.26$^\dagger$ & \na & 14.02 & \na & 31.72 & 33.11 & 31.93 & \na \\
    Merged + Attractor & 19.80$^\dagger$ & \na & 18.23 & \na & 27.40 & 19.65 & 27.38 & \na \\
    \oursrow
    E2E-SD & 16.02 & 41.93 & 1.31 & 19.71 & 2.95 & 25.29 & 18.52 & 17.05 \\
    \oursrow
    \sd{\quad $\pm$ std} & \sd{1.00} & \sd{2.60} & \sd{0.15} & \sd{1.30} & \sd{1.23} & \sd{5.73} & \sd{3.98} & \sd{6.16} \\
    \bottomrule
  \end{tabular}
  \tabnote{PS: PartialSpoof eval; LlamaPS: LlamaPartialSpoof, cross-fade part (part a); 19LA, 21LA, 21DF: ASVspoof 2019 LA, 2021 LA and DF eval; ITW: In-the-Wild (App.~\ref{app:transfer_sets}). Prior values are from the model papers, except \textsuperscript{a}Table~III of \citet{luong2025llamapartialspoof}, \textsuperscript{b}Table~III of \citet{luong2025robust} and \textsuperscript{c}Table~3 of \citet{mao2026sal} (released checkpoint). $^\dagger$Non-speech excluded, score definition not stated; E2E-SD with the same exclusion: \ms{14.25}{1.38}.}
  \end{threeparttable}
\end{table}

We evaluate the PartialSpoof-trained models on test sets outside PartialSpoof without further training (Table~\ref{tab:transfer}; App.~\ref{app:transfer_sets}). E2E-SD has no localization or detection loss; it scores each frame by its similarity to the bonafide attractor, $S_{t,0}$, and each utterance by the minimum of this score (Algorithm~\ref{alg:infer}). For localization, the seg-EER of E2E-SD on LlamaPartialSpoof, whose corpus, speakers, generators and splicing are all unseen in training, lies within the range of the localization models (41.93 vs.\ 35.52--47.49\%), while on PartialSpoof these models achieve lower seg-EERs (5.20--13.70 vs.\ 16.02\%). For detection, the utt-EER of E2E-SD on LlamaPartialSpoof is comparable to that of MRM (19.71 vs.\ 24.49\%); it rises under the codec and channel shift of ASVspoof 2021 (2.95\% on 2019 LA; 25.29 and 18.52\% on 2021 LA and DF) and is 17.05\% on In-the-Wild. All partially spoofed transfer utterances contain one spoofing method ($K{=}2$; diarization in App.~\ref{app:transfer}).

\subsection{The concatenated-part class}
\label{sec:conp}

\begin{table}[t]
  \caption{Effect of the ConP class in training (mean\,$\pm$\,std over three runs). LlamaPS $\JER$ and $K$-acc.: 32,194 partially spoofed utterances ($K{=}2$); EER: all 76,228. $\Delta$ is computed before rounding.}
  \label{tab:conp}
  \centering
  \small
  \setlength{\tabcolsep}{3.4pt}
  \begin{tabular}{lcccccc}
    \toprule
    & \multicolumn{2}{c}{PartialSpoof (in domain)} & \multicolumn{4}{c}{LlamaPS (transfer)} \\
    \cmidrule(lr){2-3}\cmidrule(l){4-7}
    Training labels & $\JER\downarrow$ & utt-EER\,$\downarrow$ & $\JER\downarrow$ & $K$-acc.\,$\uparrow$ & utt-EER\,$\downarrow$ & seg-EER\,$\downarrow$ \\
    \midrule
    without ConP & \ms{18.38}{2.42} & \ms{21.31}{7.96} & \best{\ms{30.34}{0.83}} & \best{\ms{91.77}{2.21}} & \ms{25.10}{10.19} & \ms{44.84}{5.61} \\
    \oursrow
    with ConP (default) & \best{\ms{16.01}{0.95}} & \best{\ms{1.31}{0.15}} & \ms{35.02}{2.10} & \ms{3.15}{1.55} & \best{\ms{19.71}{1.30}} & \best{\ms{41.93}{2.60}} \\
    \midrule
    \multicolumn{1}{@{}l}{\small $\Delta$ (with $-$ without)} & \gain{$-$2.37} & \gain{$-$20.00} & \loss{+4.68} & \loss{$-$88.61} & \gain{$-$5.39} & \gain{$-$2.92} \\
    \bottomrule
  \end{tabular}
\end{table}

Following the Mul-CM label set of 3C, E2E-SD keeps the concatenated parts (ConP), the non-speech regions where two segments are joined, as a training class, because they carry artifacts that help spoof detection~\citep{zhang2024spoofdia,cai2023boundary}. No other corpus we test defines ConP. Table~\ref{tab:conp} compares training with and without it.

\paragraph{Effect on class counting.}
LlamaPartialSpoof has no ConP annotation, so this labeling convention appears as a systematic overestimation by one class ($K$-acc.\ 3.15\%). Retraining without ConP raises the $K$-acc.\ to 91.77\% and lowers $\JER$ (35.02 $\to$ 30.34\%). Every partially spoofed LlamaPS utterance has $K{=}2$, so this test shows where the offset comes from; it does not test counting beyond two classes.

\paragraph{Effect on detection.}
Without ConP, the utt-EER rises from 1.31\% to 21.31\% on PartialSpoof and from 19.71\% to 25.10\% on LlamaPartialSpoof, and the in-domain $\JER$ from 16.01\% to 18.38\%. ConP is the only class that gives a target to the regions between speech segments, so it lets the model use concatenation artifacts for detection, in line with the spoofing cues found in silence~\citep{zhang2021silence,muller2021silence}. ConP is therefore a trade-off between detecting spliced audio and counting classes on a corpus without this label.

% ============================================================
\section{Analysis}
\label{sec:analysis}

\subsection{$K$ estimation reliability}
\label{sec:kest}

\begin{figure}[t]
  \centering
  \includegraphics[width=\linewidth]{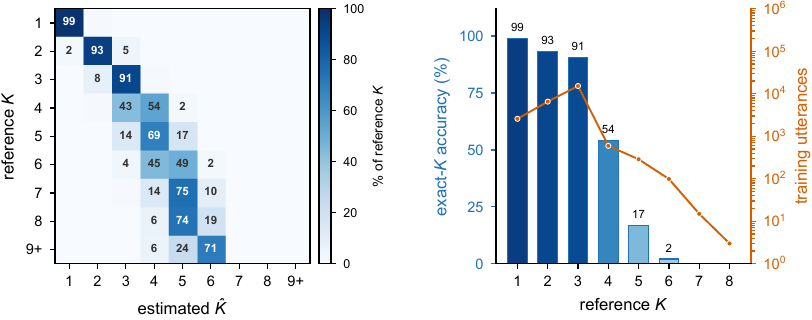}
  \caption{$K$ estimation (eval set, estimated $K$, gMLP + scaled CE, WavLM; one run). Left: ground-truth $K$ vs.\ predicted $\hat{K}$ ($K_{\max}{=}8$), row-normalized \%. Right: accuracy per $K$ (bars) and training utterances per $K$ (line, log scale).}
  \label{fig:k_est}
\end{figure}

Figure~\ref{fig:k_est} shows the $K$ estimation confusion matrix. For $K\le3$, which covers 97\% of the utterances, the accuracy is 99, 93 and 91\%. For $K\ge4$, the model mostly underestimates $K$ by one or two classes. From $K{=}3$ to $K{=}4$, the number of training utterances drops 25-fold (15,255 to 596), and the accuracy drops with it, to 54\% at $K{=}4$ and 17\% at $K{=}5$, a known limitation of EDA~\citep{horiguchi2022encoder}. When one class is lost ($\hat{K} = K{-}1$), the least-represented spoofing method is most often missed (absorbed into bonafide), not merged with a similar one: in these utterances, the bonafide cluster absorbs 33.1\% of the spoofing-method instances, and 9.6\% are merged with another method (App.~\ref{app:instances}). The 17 evaluation utterances with $K \geq 9$ (0.02\%) are systematically underestimated.

\subsection{Attractor space visualization}
\label{sec:attractors}

\begin{figure}[t]
  \centering
  \includegraphics[width=\linewidth]{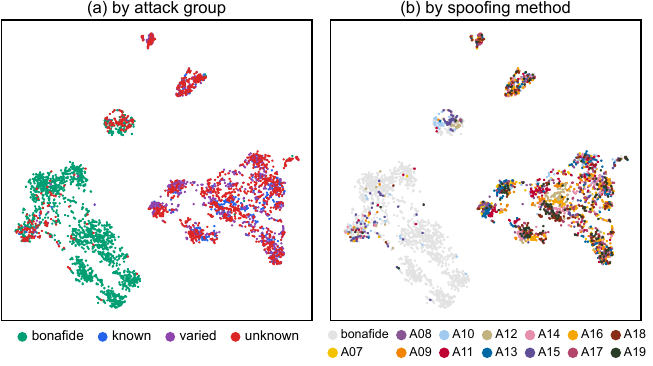}
  \caption{t-SNE visualization of learned attractors on the eval set (1,930 utterances, balanced over attacks), colored \textbf{(a)} by attack group and \textbf{(b)} by individual spoofing method.}
  \label{fig:attractors}
\end{figure}

Figure~\ref{fig:attractors} shows two properties of the learned attractor space. First, bonafide attractors ($\mathbf{a}_0$) form a single region separated from the spoof region in the t-SNE projection, consistent with the bonafide anchoring design (Section~\ref{sec:assignment}) and one-class perspectives on anti-spoofing~\citep{zhang2021ocsoftmax} that argue a well-defined bonafide boundary is more robust than attack-specific modeling.

Second, spoof attractors do not form tightly separated per-method or per-group clusters; instead, they distribute across a shared region without clear attack-type boundaries. This suggests that the EDA learns a generalized spoofing representation capturing shared artifacts rather than memorizing individual attack algorithms. Such a shared spoof region can accommodate unseen attacks without requiring dedicated cluster centroids, which may account for E2E-SD's unknown-attack generalization (Section~\ref{sec:main_results}).

% ============================================================
\section{Discussion}
\label{sec:discussion}

\paragraph{Structural advantages over modular pipelines.}
The results point to three advantages of end-to-end training:
(1)~the end-to-end model lowers $\JER$ even with the front-ends of the modular pipelines (Section~\ref{sec:main_results});
(2)~training optimizes the same scores that the argmax uses at inference, unlike training with a classification loss but evaluating with AHC;
(3)~joint $K$ estimation via existence probabilities eliminates external oracle dependency.

\paragraph{Unknown attack generalization.}
The $\JER$ reduction on unknown attacks is consistent across SSL front-ends (Section~\ref{sec:main_results}), and the t-SNE analysis suggests that it comes from a shared spoof region rather than method-specific clusters (Section~\ref{sec:attractors}). The primary failure mode is a \textit{missed method}, absorbed into bonafide rather than merged with a similar method (Section~\ref{sec:kest}).

\paragraph{Limitations.}
(1)~\textit{Single training corpus.} PartialSpoof is currently the only publicly available benchmark providing segment-level multi-class spoofing annotations, so all models are trained on it and $K_{\max}$ follows this corpus. The transfer sets contain at most one spoofing method per utterance ($K{\le}2$), so attribution between several methods outside PartialSpoof is not tested.
(2)~\textit{SSL confound.} The same-SSL comparison partially addresses the front-end confound; we did not re-run prior baselines with WavLM-Large or under our protocol.
(3)~\textit{Single speaker.} PartialSpoof has one speaker per utterance~\citep{zhang2024spoofdia}; spoof-speaker diarization with several bonafide speakers is untested, and one bonafide attractor per speaker is a natural extension.
(4)~\textit{Non-speech.} Non-speech frames have no target except ConP, so deployment requires voice activity detection.
Details are in App.~\ref{app:limitations}, and broader impacts, including the risk of false accusations, in App.~\ref{app:impacts}.

% ============================================================
\section{Conclusion}
\label{sec:conclusion}

We presented E2E-SD, a clustering-free end-to-end framework that reformulates spoof diarization as variable-cardinality single-label temporal set prediction and replaces modular clustering pipelines with encoder-decoder attractors and Hungarian bipartite matching. On the PartialSpoof benchmark, E2E-SD achieves a $\JER$ of 16.01\% while estimating the class count correctly for 88.4\% of the utterances, outperforming the best prior method by 34.3\% relative despite the baseline having access to oracle $K$. The improvement holds with the SSL front-ends of the baselines and is most pronounced on unknown attacks ($\JER$: 36.78\% $\to$ 22.33\%). More broadly, our results suggest that spoof diarization may not fundamentally require a separate clustering stage.

All models are trained on PartialSpoof, which has one speaker per utterance, and every partially spoofed transfer utterance contains a single spoofing method. Attribution between several methods outside PartialSpoof and spoof-speaker diarization with one bonafide attractor per speaker remain open, as does class counting for rare utterances with four or more classes. We hope this perspective encourages future research toward fully end-to-end forensic speech understanding systems.

\begin{ack}
This work was supported by a KOITA grant funded by MSIT (KOITA20250002-38).  This work was also supported by the Institute of Information and Communications Technology Planning and Evaluation (IITP)-ICT Challenge and Advanced Network of HRD (ICAN) Grant funded by Korea Government (MSIT) under Grant IITP-2026-RS-2024-00437024.
\end{ack}

\bibliographystyle{unsrtnat}
\bibliography{references}

@inproceedings{zhang2021initial,
  title     = {An Initial Investigation for Detecting Partially Spoofed Audio},
  author    = {Zhang, Lin and Wang, Xin and Cooper, Erica and Yamagishi, Junichi and Patino, Jose and Evans, Nicholas},
  booktitle = {Proc. Interspeech},
  pages     = {4264--4268},
  year      = {2021}
}

@article{zhang2023partialspoof,
  title   = {The {PartialSpoof} Database and Countermeasures for the Detection of Short Fake Speech Segments Embedded in an Utterance},
  author  = {Zhang, Lin and Wang, Xin and Cooper, Erica and Evans, Nicholas and Yamagishi, Junichi},
  journal = {IEEE/ACM Transactions on Audio, Speech, and Language Processing},
  volume  = {31},
  pages   = {813--825},
  year    = {2023}
}

@inproceedings{zhang2024spoofdia,
  title     = {Spoof Diarization: ``What Spoofed When'' in Partially Spoofed Audio},
  author    = {Zhang, Lin and Wang, Xin and Cooper, Erica and Diez, Mireia and Landini, Federico and Evans, Nicholas and Yamagishi, Junichi},
  booktitle = {Proc. Interspeech},
  pages     = {502--506},
  year      = {2024}
}

@inproceedings{fujita2019end,
  title     = {End-to-End Neural Speaker Diarization with Permutation-Free Objectives},
  author    = {Fujita, Yusuke and Kanda, Naoyuki and Horiguchi, Shota and Nagamatsu, Kenji and Watanabe, Shinji},
  booktitle = {Proc. Interspeech},
  pages     = {4300--4304},
  year      = {2019}
}

@inproceedings{horiguchi2020endtoend,
  title     = {End-to-End Speaker Diarization for an Unknown Number of Speakers with Encoder-Decoder Based Attractors},
  author    = {Horiguchi, Shota and Fujita, Yusuke and Watanabe, Shinji and Xue, Yawen and Nagamatsu, Kenji},
  booktitle = {Proc. Interspeech},
  pages     = {269--273},
  year      = {2020}
}

@article{horiguchi2022encoder,
  title   = {Encoder-Decoder Based Attractors for End-to-End Neural Diarization},
  author  = {Horiguchi, Shota and Fujita, Yusuke and Watanabe, Shinji and Xue, Yawen and Garc{\'i}a, Paola},
  journal = {IEEE/ACM Transactions on Audio, Speech, and Language Processing},
  volume  = {30},
  pages   = {1493--1507},
  year    = {2022}
}

@inproceedings{baevski2020wav2vec,
  title     = {wav2vec 2.0: A Framework for Self-Supervised Learning of Speech Representations},
  author    = {Baevski, Alexei and Zhou, Yuhao and Mohamed, Abdelrahman and Auli, Michael},
  booktitle = {Advances in Neural Information Processing Systems},
  volume    = {33},
  pages     = {12449--12460},
  year      = {2020}
}

@article{chen2022wavlm,
  title   = {{WavLM}: Large-Scale Self-Supervised Pre-Training for Full Stack Speech Processing},
  author  = {Chen, Sanyuan and Wang, Chengyi and Chen, Zhengyang and Wu, Yu and Liu, Shujie and Chen, Zhuo and Li, Jinyu and Kanda, Naoyuki and Yoshioka, Takuya and Xiao, Xiong and Wu, Jian and Zhou, Long and Ren, Shuo and Qian, Yanmin and Qian, Yao and Wu, Jian and Zeng, Michael and Yu, Xiangzhan and Wei, Furu},
  journal = {IEEE Journal of Selected Topics in Signal Processing},
  volume  = {16},
  number  = {6},
  pages   = {1505--1518},
  year    = {2022}
}

@article{kuhn1955hungarian,
  title   = {The {Hungarian} Method for the Assignment Problem},
  author  = {Kuhn, Harold W},
  journal = {Naval Research Logistics Quarterly},
  volume  = {2},
  number  = {1-2},
  pages   = {83--97},
  year    = {1955}
}

@inproceedings{wan2018ge2e,
  title     = {Generalized End-to-End Loss for Speaker Verification},
  author    = {Wan, Li and Wang, Quan and Papir, Alan and Moreno, Ignacio Lopez},
  booktitle = {Proc. IEEE International Conference on Acoustics, Speech and Signal Processing (ICASSP)},
  pages     = {4879--4883},
  year      = {2018}
}

@inproceedings{yu2017pit,
  title     = {Permutation Invariant Training of Deep Models for Speaker-Independent Multi-Talker Speech Separation},
  author    = {Yu, Dong and Kolb{\ae}k, Morten and Tan, Zheng-Hua and Jensen, Jesper},
  booktitle = {Proc. IEEE International Conference on Acoustics, Speech and Signal Processing (ICASSP)},
  pages     = {241--245},
  year      = {2017}
}

@inproceedings{liu2021gmlp,
  title     = {Pay Attention to {MLPs}},
  author    = {Liu, Hanxiao and Dai, Zihang and So, David and Le, Quoc V.},
  booktitle = {Advances in Neural Information Processing Systems},
  volume    = {34},
  pages     = {9204--9215},
  year      = {2021}
}

@inproceedings{tak2022automatic,
  title     = {Automatic Speaker Verification Spoofing and Deepfake Detection Using {wav2vec} 2.0 and Data Augmentation},
  author    = {Tak, Hemlata and Todisco, Massimiliano and Wang, Xin and Jung, Jee-weon and Yamagishi, Junichi and Evans, Nicholas},
  booktitle = {Proc. The Speaker and Language Recognition Workshop (Odyssey)},
  pages     = {112--119},
  year      = {2022}
}

@inproceedings{koo2025token,
  title     = {Token-based Attractors and Cross-attention in Spoof Diarization},
  author    = {Koo, Kyo-Won and Lim, Chan-yeong and Jung, Jee-weon and Shim, Hye-jin and Yu, Ha-Jin},
  booktitle = {Proc. IEEE Automatic Speech Recognition and Understanding Workshop (ASRU)},
  pages     = {1--7},
  year      = {2025}
}

@inproceedings{yi2021halfttruth,
  title     = {Half-Truth: A Partially Fake Audio Detection Dataset},
  author    = {Yi, Jiangyan and Bai, Ye and Tao, Jianhua and Ma, Haoxin and Tian, Zhengkun and Wang, Chenglong and Wang, Tao and Fu, Ruibo},
  booktitle = {Proc. Interspeech},
  pages     = {1654--1658},
  year      = {2021}
}

@inproceedings{wu2022fakesspan,
  title     = {Partially Fake Audio Detection by Self-Attention-Based Fake Span Discovery},
  author    = {Wu, Haibin and Kuo, Heng-Cheng and Zheng, Naijun and Hung, Kuo-Hsuan and Lee, Hung-Yi and Tsao, Yu and Wang, Hsin-Min and Meng, Helen},
  booktitle = {Proc. IEEE International Conference on Acoustics, Speech and Signal Processing (ICASSP)},
  pages     = {9236--9240},
  year      = {2022}
}

@inproceedings{cai2023boundary,
  title     = {Waveform Boundary Detection for Partially Spoofed Audio},
  author    = {Cai, Zexin and Wang, Weiqing and Li, Ming},
  booktitle = {Proc. IEEE International Conference on Acoustics, Speech and Signal Processing (ICASSP)},
  pages     = {1--5},
  year      = {2023}
}

@inproceedings{zhang2023rangeeer,
  title     = {Range-Based Equal Error Rate for Spoof Localization},
  author    = {Zhang, Lin and Wang, Xin and Cooper, Erica and Evans, Nicholas and Yamagishi, Junichi},
  booktitle = {Proc. Interspeech},
  pages     = {3212--3216},
  year      = {2023}
}

@inproceedings{jung2022aasist,
  title     = {{AASIST}: Audio Anti-Spoofing Using Integrated Spectro-Temporal Graph Attention Networks},
  author    = {Jung, Jee-weon and Heo, Hee-Soo and Tak, Hemlata and Shim, Hye-jin and Chung, Joon Son and Lee, Bong-Jin and Yu, Ha-Jin and Evans, Nicholas},
  booktitle = {Proc. IEEE International Conference on Acoustics, Speech and Signal Processing (ICASSP)},
  pages     = {6367--6371},
  year      = {2022}
}

@inproceedings{tak2021rawnet2,
  title     = {End-to-End Anti-Spoofing with {RawNet2}},
  author    = {Tak, Hemlata and Patino, Jose and Todisco, Massimiliano and Nautsch, Andreas and Evans, Nicholas and Larcher, Anthony},
  booktitle = {Proc. IEEE International Conference on Acoustics, Speech and Signal Processing (ICASSP)},
  pages     = {6369--6373},
  year      = {2021}
}

@misc{yamagishi2019asvspoof,
  title        = {{ASVspoof} 2019: The 3rd Automatic Speaker Verification Spoofing and Countermeasures Challenge Database},
  author       = {Yamagishi, Junichi and Todisco, Massimiliano and Sahidullah, Md and Delgado, H{\'e}ctor and Wang, Xin and Evans, Nicholas and Kinnunen, Tomi and Lee, Kong Aik and Vestman, Ville and Nautsch, Andreas},
  howpublished = {University of Edinburgh, The Centre for Speech Technology Research (CSTR)},
  year         = {2019}
}

@inproceedings{carion2020detr,
  title     = {End-to-End Object Detection with Transformers},
  author    = {Carion, Nicolas and Massa, Francisco and Synnaeve, Gabriel and Usunier, Nicolas and Kirillov, Alexander and Zagoruyko, Sergey},
  booktitle = {Proc. European Conference on Computer Vision (ECCV)},
  pages     = {213--229},
  year      = {2020}
}

@article{park2022review,
  title   = {A Review of Speaker Diarization: Recent Advances with Deep Learning},
  author  = {Park, Tae Jin and Kanda, Naoyuki and Dimitriadis, Dimitrios and Han, Kyu J and Watanabe, Shinji and Narayanan, Shrikanth},
  journal = {Computer Speech \& Language},
  volume  = {72},
  pages   = {101317},
  year    = {2022}
}

@article{wang2023valle,
  title   = {Neural Codec Language Models are Zero-Shot Text to Speech Synthesizers},
  author  = {Chen, Sanyuan and Wang, Chengyi and Wu, Yu and Zhang, Ziqiang and Zhou, Long and Liu, Shujie and Chen, Zhuo and Liu, Yanqing and Wang, Huaming and Li, Jinyu and He, Lei and Zhao, Sheng and Wei, Furu},
  journal = {IEEE Transactions on Audio, Speech and Language Processing},
  volume  = {33},
  pages   = {705--718},
  year    = {2025}
}

@inproceedings{yi2022add,
  title     = {{ADD} 2022: The First Audio Deep Synthesis Detection Challenge},
  author    = {Yi, Jiangyan and Fu, Ruibo and Tao, Jianhua and Nie, Shuai and Ma, Haoxin and Wang, Chenglong and Wang, Tao and Tian, Zhengkun and Bai, Ye and Fan, Cunhang and Liang, Shan and Wang, Shiming and Zhang, Shuai and Yan, Xinrui and Xu, Le and Wen, Zhengqi and Li, Haizhou},
  booktitle = {Proc. IEEE International Conference on Acoustics, Speech and Signal Processing (ICASSP)},
  pages     = {9216--9220},
  year      = {2022}
}

@inproceedings{horiguchi2021global,
  title     = {Towards Neural Diarization for Unlimited Numbers of Speakers Using Global and Local Attractors},
  author    = {Horiguchi, Shota and Watanabe, Shinji and Garc{\'i}a, Paola and Xue, Yawen and Takashima, Yuki and Kawaguchi, Yohei},
  booktitle = {Proc. IEEE Automatic Speech Recognition and Understanding Workshop (ASRU)},
  pages     = {98--105},
  year      = {2021}
}

@inproceedings{samarakoon2023transformer,
  title     = {Transformer Attractors for Robust and Efficient End-to-End Neural Diarization},
  author    = {Samarakoon, Lahiru and Broughton, Samuel J. and H{\"a}rk{\"o}nen, Marc and Fung, Ivan},
  booktitle = {Proc. IEEE Automatic Speech Recognition and Understanding Workshop (ASRU)},
  pages     = {1--8},
  year      = {2023}
}

@inproceedings{chen2023aed,
  title     = {Attention-based Encoder-Decoder Network for End-to-End Neural Speaker Diarization with Target Speaker Attractor},
  author    = {Chen, Zhengyang and Han, Bing and Wang, Shuai and Qian, Yanmin},
  booktitle = {Proc. Interspeech},
  pages     = {3552--3556},
  year      = {2023}
}

@inproceedings{liu2024spoofcm,
  title     = {How Do Neural Spoofing Countermeasures Detect Partially Spoofed Audio?},
  author    = {Liu, Tianchi and Zhang, Lin and Das, Rohan Kumar and Ma, Yi and Tao, Ruijie and Li, Haizhou},
  booktitle = {Proc. Interspeech},
  pages     = {1105--1109},
  year      = {2024}
}

@inproceedings{zhong2024boundary,
  title     = {Enhancing Partially Spoofed Audio Localization with Boundary-Aware Attention Mechanism},
  author    = {Zhong, Jiafeng and Li, Bin and Yi, Jiangyan},
  booktitle = {Proc. Interspeech},
  pages     = {4838--4842},
  year      = {2024}
}

@inproceedings{zhang2019p2sgrad,
  title     = {{P2SGrad}: Refined Gradients for Optimizing Deep Face Models},
  author    = {Zhang, Xiao and Zhao, Rui and Yan, Junjie and Gao, Mengya and Qiao, Yu and Wang, Xiaogang and Li, Hongsheng},
  booktitle = {Proc. IEEE/CVF Conference on Computer Vision and Pattern Recognition (CVPR)},
  pages     = {9898--9906},
  year      = {2019}
}

@inproceedings{mueller2022generalize,
  title     = {Does Audio Deepfake Detection Generalize?},
  author    = {M{\"u}ller, Nicolas M. and Czempin, Pavel and Dieckmann, Franziska and Froghyar, Adam and B{\"o}ttinger, Konstantin},
  booktitle = {Proc. Interspeech},
  pages     = {2783--2787},
  year      = {2022}
}

@inproceedings{shih2024artifacts,
  title     = {Does Audio Deepfake Detection Rely on Artifacts?},
  author    = {Shih, Tsu-Hsien and Yeh, Chin-Yuan and Chen, Ming-Syan},
  booktitle = {Proc. IEEE International Conference on Acoustics, Speech and Signal Processing (ICASSP)},
  pages     = {12446--12450},
  year      = {2024}
}

@article{yi2023addsurvey,
  title   = {Audio Deepfake Detection: A Survey},
  author  = {Yi, Jiangyan and Wang, Chenglong and Tao, Jianhua and Zhang, Xiaohui and Zhang, Chu Yuan and Zhao, Yan},
  journal = {arXiv preprint arXiv:2308.14970},
  year    = {2023}
}

@inproceedings{gulati2020conformer,
  title     = {Conformer: Convolution-augmented Transformer for Speech Recognition},
  author    = {Gulati, Anmol and Qin, James and Chiu, Chung-Cheng and Parmar, Niki and Zhang, Yu and Yu, Jiahui and Han, Wei and Wang, Shibo and Zhang, Zhengdong and Wu, Yonghui and Pang, Ruoming},
  booktitle = {Proc. Interspeech},
  pages     = {5036--5040},
  year      = {2020}
}

@inproceedings{vaswani2017attention,
  title     = {Attention Is All You Need},
  author    = {Vaswani, Ashish and Shazeer, Noam and Parmar, Niki and Uszkoreit, Jakob and Jones, Llion and Gomez, Aidan N. and Kaiser, {\L}ukasz and Polosukhin, Illia},
  booktitle = {Advances in Neural Information Processing Systems},
  volume    = {30},
  pages     = {5998--6008},
  year      = {2017}
}

@inproceedings{snyder2018xvector,
  title     = {X-Vectors: Robust {DNN} Embeddings for Speaker Recognition},
  author    = {Snyder, David and Garcia-Romero, Daniel and Sell, Gregory and Povey, Daniel and Khudanpur, Sanjeev},
  booktitle = {Proc. IEEE International Conference on Acoustics, Speech and Signal Processing (ICASSP)},
  pages     = {5329--5333},
  year      = {2018}
}

@inproceedings{wu2015asvspoof,
  title     = {{ASVspoof} 2015: The First Automatic Speaker Verification Spoofing and Countermeasures Challenge},
  author    = {Wu, Zhizheng and Kinnunen, Tomi and Evans, Nicholas and Yamagishi, Junichi and Hanil{\c{c}}i, Cemal and Sahidullah, Md and Sizov, Aleksandr},
  booktitle = {Proc. Interspeech},
  pages     = {2037--2041},
  year      = {2015}
}

@inproceedings{kinnunen2017asvspoof,
  title     = {The {ASVspoof} 2017 Challenge: Assessing the Limits of Replay Spoofing Attack Detection},
  author    = {Kinnunen, Tomi and Sahidullah, Md and Delgado, H{\'e}ctor and Todisco, Massimiliano and Evans, Nicholas and Yamagishi, Junichi and Lee, Kong Aik},
  booktitle = {Proc. Interspeech},
  pages     = {2--6},
  year      = {2017}
}

@inproceedings{todisco2019asvspoof,
  title     = {{ASVspoof} 2019: Future Horizons in Spoofed and Fake Audio Detection},
  author    = {Todisco, Massimiliano and Wang, Xin and Vestman, Ville and Sahidullah, Md and Delgado, H{\'e}ctor and Nautsch, Andreas and Yamagishi, Junichi and Evans, Nicholas and Kinnunen, Tomi H. and Lee, Kong Aik},
  booktitle = {Proc. Interspeech},
  pages     = {1008--1012},
  year      = {2019}
}

@inproceedings{yamagishi2021asvspoof,
  title     = {{ASVspoof} 2021: Accelerating Progress in Spoofed and Deepfake Speech Detection},
  author    = {Yamagishi, Junichi and Wang, Xin and Todisco, Massimiliano and Sahidullah, Md and Patino, Jose and Nautsch, Andreas and Liu, Xuechen and Lee, Kong Aik and Kinnunen, Tomi and Evans, Nicholas and Delgado, H{\'e}ctor},
  booktitle = {Proc. 2021 Edition of the Automatic Speaker Verification and Spoofing Countermeasures Challenge},
  pages     = {47--54},
  year      = {2021}
}

@inproceedings{fujita2019saend,
  title     = {End-to-End Neural Speaker Diarization with Self-Attention},
  author    = {Fujita, Yusuke and Kanda, Naoyuki and Horiguchi, Shota and Xue, Yawen and Nagamatsu, Kenji and Watanabe, Shinji},
  booktitle = {Proc. IEEE Automatic Speech Recognition and Understanding Workshop (ASRU)},
  pages     = {296--303},
  year      = {2019}
}

@inproceedings{babu2022xlsr,
  title     = {{XLS-R}: Self-supervised Cross-lingual Speech Representation Learning at Scale},
  author    = {Babu, Arun and Wang, Changhan and Tjandra, Andros and Lakhotia, Kushal and Xu, Qiantong and Goyal, Naman and Singh, Kritika and von Platen, Patrick and Saraf, Yatharth and Pino, Juan and Baevski, Alexei and Conneau, Alexis and Auli, Michael},
  booktitle = {Proc. Interspeech},
  pages     = {2278--2282},
  year      = {2022}
}

@inproceedings{kingma2015adam,
  title     = {Adam: A Method for Stochastic Optimization},
  author    = {Kingma, Diederik P and Ba, Jimmy},
  booktitle = {Proc. International Conference on Learning Representations (ICLR)},
  year      = {2015}
}

@inproceedings{yi2023add2,
  title     = {{ADD} 2023: The Second Audio Deepfake Detection Challenge},
  author    = {Yi, Jiangyan and Tao, Jianhua and Fu, Ruibo and Yan, Xinrui and Wang, Chenglong and Wang, Tao and Zhang, Chu Yuan and Zhang, Xiaohui and Zhao, Yan and Ren, Yong and Xu, Le and Zhou, Junzuo and Gu, Hao and Wen, Zhengqi and Liang, Shan and Lian, Zheng and Nie, Shuai and Li, Haizhou},
  booktitle = {Proc. IJCAI 2023 Workshop on Deepfake Audio Detection and Analysis (DADA)},
  series    = {CEUR Workshop Proceedings},
  volume    = {3597},
  pages     = {125--130},
  year      = {2023}
}

@inproceedings{kinoshita2021integrating,
  title     = {Integrating End-to-End Neural and Clustering-Based Diarization: Getting the Best of Both Worlds},
  author    = {Kinoshita, Keisuke and Delcroix, Marc and Tawara, Naohiro},
  booktitle = {Proc. IEEE International Conference on Acoustics, Speech and Signal Processing (ICASSP)},
  pages     = {7198--7202},
  year      = {2021}
}

@article{dehak2011ivector,
  title   = {Front-End Factor Analysis for Speaker Verification},
  author  = {Dehak, Najim and Kenny, Patrick J. and Dehak, R{\'e}da and Dumouchel, Pierre and Ouellet, Pierre},
  journal = {IEEE Transactions on Audio, Speech, and Language Processing},
  volume  = {19},
  number  = {4},
  pages   = {788--798},
  year    = {2011}
}

@inproceedings{muller2021silence,
  title     = {Speech is Silver, Silence is Golden: What do {ASVspoof}-trained Models Really Learn?},
  author    = {M{\"u}ller, Nicolas M. and Dieckmann, Franziska and Czempin, Pavel and Canals, Roman and B{\"o}ttinger, Konstantin and Williams, Jennifer},
  booktitle = {Proc. 2021 Edition of the Automatic Speaker Verification and Spoofing Countermeasures Challenge},
  pages     = {55--60},
  year      = {2021}
}

@inproceedings{zhang2021silence,
  title     = {The Effect of Silence and Dual-Band Fusion in Anti-Spoofing System},
  author    = {Zhang, Yuxiang and Wang, Wenchao and Zhang, Pengyuan},
  booktitle = {Proc. Interspeech},
  pages     = {4279--4283},
  year      = {2021}
}

@inproceedings{luong2025llamapartialspoof,
  title     = {{LlamaPartialSpoof}: An {LLM}-Driven Fake Speech Dataset Simulating Disinformation Generation},
  author    = {Luong, Hieu-Thi and Li, Haoyang and Zhang, Lin and Lee, Kong Aik and Chng, Eng Siong},
  booktitle = {Proc. IEEE International Conference on Acoustics, Speech and Signal Processing (ICASSP)},
  pages     = {1--5},
  year      = {2025}
}

@inproceedings{luong2025robust,
  title     = {Robust Localization of Partially Fake Speech: Metrics and Out-of-Domain Evaluation},
  author    = {Luong, Hieu-Thi and Rimon, Inbal and Permuter, Haim and Lee, Kong Aik and Chng, Eng Siong},
  booktitle = {Proc. Asia Pacific Signal and Information Processing Association Annual Summit and Conference (APSIPA ASC)},
  pages     = {2205--2210},
  year      = {2025}
}

@inproceedings{zhang2025partialedit,
  title     = {{PartialEdit}: Identifying Partial Deepfakes in the Era of Neural Speech Editing},
  author    = {Zhang, You and Tian, Baotong and Zhang, Lin and Duan, Zhiyao},
  booktitle = {Proc. Interspeech},
  pages     = {5353--5357},
  year      = {2025}
}

@inproceedings{wu2024cfprf,
  title     = {Coarse-to-Fine Proposal Refinement Framework for Audio Temporal Forgery Detection and Localization},
  author    = {Wu, Junyan and Lu, Wei and Luo, Xiangyang and Yang, Rui and Wang, Qian and Cao, Xiaochun},
  booktitle = {Proc. ACM International Conference on Multimedia (MM)},
  pages     = {7395--7403},
  year      = {2024}
}

@inproceedings{mao2026sal,
  title     = {Localizing Speech Deepfakes Beyond Transitions via Segment-Aware Learning},
  author    = {Mao, Yuchen and Huang, Wen and Qian, Yanmin},
  booktitle = {Proc. IEEE International Conference on Acoustics, Speech and Signal Processing (ICASSP)},
  pages     = {16507--16511},
  year      = {2026}
}

@inproceedings{kim2026transition,
  title     = {Temporal Transition-Aware Multi-Head Modeling for Partially Spoofed Audio Detection and Localization},
  author    = {Kim, Yunsu and Lee, Juyeob and Park, Eunil},
  booktitle = {Proc. Interspeech},
  pages     = {3062--3070},
  year      = {2026}
}

@misc{dscore,
  title        = {dscore: Diarization Scoring Tools},
  author       = {Ryant, Neville},
  howpublished = {\url{https://github.com/nryant/dscore}},
  note         = {Commit e02f949, BSD 2-Clause License},
  year         = {2018}
}

@inproceedings{muller2022attacker,
  title     = {Attacker Attribution of Audio Deepfakes},
  author    = {M{\"u}ller, Nicolas M. and Dieckmann, Franziska and Williams, Jennifer},
  booktitle = {Proc. Interspeech},
  pages     = {2788--2792},
  year      = {2022}
}

@inproceedings{klein2024source,
  title     = {Source Tracing of Audio Deepfake Systems},
  author    = {Klein, Nicholas and Chen, Tianxiang and Tak, Hemlata and Casal, Ricardo and Khoury, Elie},
  booktitle = {Proc. Interspeech},
  pages     = {1100--1104},
  year      = {2024}
}

@inproceedings{klein2025openset,
  title     = {Open-Set Source Tracing of Audio Deepfake Systems},
  author    = {Klein, Nicholas and Tak, Hemlata and Khoury, Elie},
  booktitle = {Proc. Interspeech},
  pages     = {1578--1582},
  year      = {2025}
}

@article{landini2024diaper,
  title   = {{DiaPer}: End-to-End Neural Diarization With Perceiver-Based Attractors},
  author  = {Landini, Federico and Diez, Mireia and Stafylakis, Themos and Burget, Luk{\'a}{\v{s}}},
  journal = {IEEE/ACM Transactions on Audio, Speech, and Language Processing},
  volume  = {32},
  pages   = {3450--3465},
  year    = {2024}
}

@article{zhang2021ocsoftmax,
  title   = {One-Class Learning Towards Synthetic Voice Spoofing Detection},
  author  = {Zhang, You and Jiang, Fei and Duan, Zhiyao},
  journal = {IEEE Signal Processing Letters},
  volume  = {28},
  pages   = {937--941},
  year    = {2021}
}

@inproceedings{elkheir2025comprehensive,
  title     = {Comprehensive Layer-wise Analysis of {SSL} Models for Audio Deepfake Detection},
  author    = {El Kheir, Yassine and Samih, Younes and Maharjan, Suraj and Polzehl, Tim and M{\"o}ller, Sebastian},
  booktitle = {Findings of the Association for Computational Linguistics: NAACL 2025},
  pages     = {4070--4082},
  year      = {2025}
}

%%%%%%%%%%%%%%%%%%%%%%%%%%%%%%%%%%%%%%%%%%%%%%%%%%%%%%%%%%%%

\clearpage
\appendix
\raggedbottom
% Appendix title page with its own table of contents (titletoc, prefix "p")
\noindent{\fontsize{28}{32}\selectfont\bfseries Appendix}\par
\vspace{2.2em}
\noindent{\LARGE\bfseries Table of Contents}\par
\vspace{0.4em}
\noindent\rule{\linewidth}{0.6pt}
\startcontents[appendix]
{\setlength{\parskip}{0pt}\printcontents[appendix]{p}{1}{\setcounter{tocdepth}{2}}}
\vspace{1.2em}
\noindent\rule{\linewidth}{0.6pt}
\clearpage

\section{Implementation Details}
\label{app:impl}

\subsection{Architecture and Training Configuration}
\label{app:config}

Table~\ref{tab:hparams} gives the configuration of every run. An ablation changes one factor and keeps all others.
The SSL front-end is fine-tuned with the same base learning rate as the rest of the network.
Early stopping uses the development loss $\Ltotal$.
The checkpoint with the lowest development loss is evaluated.
Each utterance is processed at full length, without cropping, windowing or data augmentation.
Mini-batches are zero-padded to the longest utterance, and padded frames have no target.

\paragraph{Number of attractors.}
During training the decoder generates $\max_i K_i + 1$ attractors for a mini-batch, where $K_i$ is the number of classes of utterance $i$.
The attractor at index $K_i$ is the stop attractor of utterance $i$ (existence target 0).
Attractors after it receive existence target 0 and no frames.
The budget $\Kmax$ therefore does not change training. It only limits $\hat K$ at inference.
We set $\Kmax$ to the largest number of co-occurring classes in the training and development labels (8, Table~\ref{tab:classes}).
For another corpus, $\Kmax$ must be set again by the same rule.

\paragraph{Scale of the cosine logits.}
In Eq.~\ref{eq:scaled_sim}, $w$ and $b$ are initialized to 10 and $-5$ and updated with $0.01\times$ the base learning rate.

\begin{table}[H]
  \caption{Configuration of E2E-SD (all runs unless an ablation changes one factor).}
  \label{tab:hparams}
  \centering\small
  \begin{tabular}{@{}lll@{}}
    \toprule
    Group & Item & Value \\
    \midrule
    \multirow{3}{*}{Encoder}
      & SSL front-end & WavLM-Large, fine-tuned; softmax-weighted sum of all hidden states \\
      & Back-end & gMLP, 5 layers, $d{=}512$, dropout 0.1 on SSL features \\
      & Frame rate & 20\,ms \\
    \midrule
    \multirow{3}{*}{EDA}
      & Encoder / decoder & 1-layer LSTM each, hidden size 512 \\
      & Time shuffle & on (training and inference) \\
      & Existence head & linear, sigmoid; unweighted BCE, $\lambda{=}1.0$ \\
    \midrule
    \multirow{2}{*}{Assignment}
      & Similarity & cosine + affine $wS{+}b$, init.\ $(10,-5)$ (Eq.~\ref{eq:scaled_sim}) \\
      & Matching & $\mathbf{a}_0$ fixed to bonafide; Hungarian over spoof classes \\
    \midrule
    \multirow{6}{*}{Optimization}
      & Optimizer & Adam~\tcite{kingma2015adam} ($\beta_1{=}0.9$, $\beta_2{=}0.999$), weight decay $10^{-6}$ \\
      & Learning rate & $10^{-5}$ incl.\ SSL ($w$, $b$: $0.01\times$, no decay); halved every 10 ep. \\
      & Batch & 2 utterances $\times$ 4 accumulation steps (effective 8), full length \\
      & Gradient clipping & norm 3.0 \\
      & Stopping & dev loss, patience 5, at most 100 epochs \\
      & Augmentation & none \\
    \midrule
    \multirow{2}{*}{Inference}
      & Class count & $\hat K = \min\{k: p_k < \theta\}$, $\theta{=}0.5$, $\Kmax{=}8$ \\
      & Frame label & $\argmax_k$ over the first $\hat K$ attractors \\
    \bottomrule
  \end{tabular}
\end{table}

\subsection{Training and Inference Algorithms}
\label{app:algorithms}

Algorithm~\ref{alg:train} gives one training step for one utterance, and Algorithm~\ref{alg:infer} gives inference.
Each line refers to its equation in Section~\ref{sec:method}.
The full implementation, with instructions for data preparation, training, inference and evaluation, is publicly available at the address below, and the trained default model is linked from the same address.

\noindent\url{https://github.com/Bongsujung/E2E-SD}

\begin{algorithm}[H]
  \caption{Training step of E2E-SD for one utterance}
  \label{alg:train}
  \begin{algorithmic}[1]
    \Require waveform $\mathbf{x}$; frame labels $c_{1:T}$ over $K$ classes (class 0: bonafide)
    \State $\mathbf{E} \gets$ frame embeddings of $\mathbf{x}$ \Comment{Sec.~\ref{sec:encoder}}
    \State $\mathbf{a}_{0:K},\, p_{0:K} \gets \mathrm{EDA}$ on the time-shuffled $\mathbf{E}$ \Comment{Eqs.~\eqref{eq:eda_enc}--\eqref{eq:exist}}
    \State $S \gets$ cosine similarity of each frame to $\mathbf{a}_{0:K-1}$ \Comment{Eq.~\eqref{eq:cos_sim}}
    \State $\pi^* \gets$ Hungarian matching of classes to attractors, with $\pi^*(0){=}0$ \Comment{Eq.~\eqref{eq:hungarian}}
    \State $\Lassign \gets$ cross-entropy of $\hat S = wS + b$ with target $\pi^*(c_t)$ \Comment{Eqs.~\eqref{eq:scaled_sim}, \eqref{eq:assign}}
    \State $\Lexist \gets$ BCE of $p_{0:K}$ with target $(1,\dots,1,0)$ \Comment{Eq.~\eqref{eq:exist}}
    \State \Return $\Lassign + \lambda\Lexist$
  \end{algorithmic}
\end{algorithm}

\begin{algorithm}[H]
  \caption{Inference of E2E-SD}
  \label{alg:infer}
  \begin{algorithmic}[1]
    \Require waveform $\mathbf{x}$; threshold $\theta$; maximum number of attractors $\Kmax$
    \State $\mathbf{E} \gets$ frame embeddings of $\mathbf{x}$
    \State $\mathbf{a}_{0:\Kmax-1},\, p_{0:\Kmax-1} \gets \mathrm{EDA}$ on the time-shuffled $\mathbf{E}$
    \State $\hat K \gets$ first $k$ with $p_k < \theta$ ($\Kmax$ if none) \Comment{class count, Eq.~\eqref{eq:auto_k}}
    \State $\hat c_t \gets \argmax_{k < \hat K} \hat S_{t,k}$ for each frame $t$ \Comment{diarization, Eq.~\eqref{eq:auto_k}}
    \State frame score $S_{t,0}$; utterance score $\min_t S_{t,0}$ \Comment{localization, detection}
    \State \Return $\hat K$, $\hat c_{1:T}$, frame scores, utterance score
  \end{algorithmic}
\end{algorithm}

\subsection{Model Size and Computational Cost}
\label{app:compute}

Table~\ref{tab:params} gives the number of parameters.

\begin{table}[H]
  \caption{Parameters used in the forward pass (millions). Task-specific = back-end + EDA + heads.}
  \label{tab:params}
  \centering\small
  \begin{threeparttable}
  \begin{tabular}{lrrrrr}
    \toprule
    Configuration & SSL & Back-end & EDA & Total & Task-specific \\
    \midrule
    \oursrow
    WavLM-L + gMLP (default)       & 315.45 & 24.51 & 4.20 & 344.16 & 28.71 \\
    XLS-R 300M / w2v2-L + gMLP     & 317.39 & 24.51 & 4.20 & 346.10 & 28.71 \\
    WavLM-L + Transformer          & 315.45 & 16.29 & 4.20 & 335.94 & 20.49 \\
    WavLM-L + Conformer            & 315.45 & 30.83 & 4.20 & 350.48 & 35.03 \\
    \bottomrule
  \end{tabular}
  \tabnote{The gMLP code inherited from the PartialSpoof recipe also builds an utterance-level branch (24.51M) that the forward pass does not use. This branch is excluded here. The three back-ends do not have matched capacity.}
  \end{threeparttable}
\end{table}

The task-specific part (back-end, EDA and heads) is 28.71M, 8.3\% of the 344.16M parameters.
3C needs two such models, one for each branch.

All runs use one NVIDIA RTX 4090 (24\,GB).
Inference on the 71,237 PartialSpoof evaluation utterances takes less than one hour.
The experiments in this paper use 60 training runs (about 340 GPU-hours) and about 100 GPU-hours of inference on the transfer and detection sets.
Preliminary experiments with an earlier version of the code used 41 further training runs (about 240 GPU-hours).

\section{Datasets and Evaluation Protocol}
\label{app:data}

\subsection{PartialSpoof}
\label{app:partialspoof}

PartialSpoof~\citep{zhang2021initial,zhang2023partialspoof} is built from ASVspoof 2019 LA~\citep{yamagishi2019asvspoof,todisco2019asvspoof}.
Each utterance concatenates bonafide segments and segments from one or more spoofing methods.
The labels have 20\,ms resolution and come in three kinds: bonafide, one class per spoofing method, and the concatenated parts (ConP).
ConP marks the regions where two segments are joined.
Table~\ref{tab:classes} gives the number of classes per utterance and the attack groups.
We use the official splits and the v1.3 voice activity labels (\texttt{vad\_20sil}).
Non-speech frames other than ConP have no training target and are not scored.
We train on the training set only and use no other corpus.

\begin{table}[H]
  \caption{PartialSpoof: number of utterances by the number of classes $K$ per utterance (bonafide, spoofing methods and ConP). The largest number of co-occurring classes in train and dev is 8, which sets $\Kmax{=}8$.}
  \label{tab:classes}
  \centering\small
  \setlength{\tabcolsep}{4.5pt}
  \begin{tabular}{@{}lrrrrrrrrrr@{}}
    \toprule
    Split & $K{=}1$ & 2 & 3 & 4 & 5 & 6 & 7 & 8 & 9--11 & Total \\
    \midrule
    Train & 2,580 & 6,544 & 15,255 & 596 & 288 & 99 & 15 & 3 & 0 & 25,380 \\
    Dev   & 2,548 & 6,639 & 15,061 & 365 & 171 & 44 & 15 & 1 & 0 & 24,844 \\
    Eval  & 7,355 & 16,602 & 44,948 & 1,236 & 662 & 284 & 102 & 31 & 17 & 71,237 \\
    \bottomrule
  \end{tabular}

  \vspace{0.6em}
  \begin{threeparttable}
  \begin{tabular}{@{}lll@{}}
    \toprule
    Group & Attacks & Definition~\citep{zhang2024spoofdia} \\
    \midrule
    \Kn{Known}   & A16, A19 & identical to a training attack \\
    \Va{Varied}  & A07, A08, A09, A17 & acoustic model or waveform generator seen in training \\
    \Un{Unknown} & A10--A15, A18 & neither seen in training \\
    \bottomrule
  \end{tabular}
  \tabnote{Training and development sets use attacks A01--A06; the evaluation set uses A07--A19. 2,153 evaluation utterances contain no bonafide speech.}
  \end{threeparttable}
\end{table}

\subsection{Evaluation Metrics}
\label{app:metrics}

\paragraph{Notation.}
Let $\mathcal{D}$ be the set of evaluation utterances.
Utterance $j$ has reference classes $\{0,\dots,K_j{-}1\}$ ($0$ = bonafide if present).
Here $\mathcal{R}^{(j)}_c$ is the set of reference intervals of class $c$, and $\mathcal{H}^{(j)}_c$ is the set of hypothesis intervals mapped to $c$.
The mapping is the Hungarian assignment that minimizes the total Jaccard error between reference classes and hypothesis clusters, as in speaker diarization scoring.
This mapping is independent of the training-time matching (Eq.~\ref{eq:hungarian}).

\paragraph{Per-class Jaccard error.}
\begin{equation}
  J^{(j)}_c = \frac{\mathrm{FA}^{(j)}_c + \mathrm{Miss}^{(j)}_c}{\big|\mathcal{R}^{(j)}_c \cup \mathcal{H}^{(j)}_c\big|},
  \qquad
  \mathrm{FA}^{(j)}_c = \big|\mathcal{H}^{(j)}_c \setminus \mathcal{R}^{(j)}_c\big|,\quad
  \mathrm{Miss}^{(j)}_c = \big|\mathcal{R}^{(j)}_c \setminus \mathcal{H}^{(j)}_c\big|,
\end{equation}
where $|\cdot|$ is total duration inside the scored region (reference speech).

\paragraph{$\JI$ and $\JER$.}
\begin{equation}
  \JI = \frac{1}{|\mathcal{D}_{\text{bona}}|} \sum_{j \in \mathcal{D}_{\text{bona}}} J^{(j)}_0,
  \qquad
  \JER = \frac{1}{\sum_{j \in \mathcal{D}} (K_j - 1)} \sum_{j \in \mathcal{D}} \sum_{c=1}^{K_j-1} J^{(j)}_c,
  \label{eq:jer}
\end{equation}
where $\mathcal{D}_{\text{bona}}$ contains the 69,084 evaluation utterances that have bonafide speech.
The bonafide error is not defined for the other 2,153 utterances.
3C and \citet{koo2025token} average over all files without stating how such files are scored.
Their $\JI$ is therefore only approximately comparable to ours.
If we include these 2,153 utterances with $J^{(j)}_0{=}0$ or with $J^{(j)}_0{=}1$, our $\JI$ becomes 7.53 or 10.56.
Both values are below the lowest prior $\JI$ in Table~\ref{tab:main} (15.15).
The $\JER$ is pooled over (utterance, method) pairs, as in Eq.~4 of 3C.
The reference contains only bonafide and spoofing-method segments, so ConP and non-speech are not scored.
As in dscore, a reference class without a mapped cluster has $J^{(j)}_c{=}1$, and a hypothesis cluster without a mapped class adds no term of its own.
Its frames still count as misses of the reference classes they cover.
Per-method and per-group $\JER$ pool the same terms over the utterances that contain the method or a method of the group.

\paragraph{Sensitivity to time stamps.}
Each $J^{(j)}_c$ is a ratio of durations.
A boundary placed $\delta$ seconds away from the reference adds $\delta$ to the false alarm or the miss of the two classes that meet at the boundary.
The error therefore grows with the boundary offset and is larger for short segments.
A wrong class count has a different cost: a missed spoofing method gives $J^{(j)}_c = 1$ for that term.

\paragraph{Implementation.}
We compute $J^{(j)}_c$ with dscore~\citep{dscore}, with the reference speech as the scoring region.
The step is 0.1\,ms; the default 10\,ms step fails on classes shorter than 10\,ms.

\paragraph{Localization and detection EER.}
The seg-EER is the equal error rate over 20\,ms frames.
Bonafide is the target class, and all frames are scored.
PartialSpoof uses its official 20\,ms segment labels. For LlamaPS and PartialEdit, a frame that overlaps a spoofed region by any amount is labeled spoof.
The utt-EER is the equal error rate over utterances.
For E2E-SD, the frame score is $\cos(\mathbf{e}_t, \mathbf{a}_0)$ and the utterance score is its minimum over frames (Algorithm~\ref{alg:infer}).

\subsection{Protocol Comparison with Prior Spoof Diarization Work}
\label{app:protocol}

Table~\ref{tab:protocol} compares the protocols.
We did not retrain the baselines.
Their numbers are taken from the original papers.
The 3C $\JI$ and $\JER$ are from Table~1 of \citet{zhang2024spoofdia}.
The other 3C numbers are from Appendix Table~2 of its arXiv version, which the paper cites for the per-method breakdown.

\begin{table}[H]
  \caption{Protocol comparison with prior spoof diarization work, as stated in the papers.}
  \label{tab:protocol}
  \centering\small
  \setlength{\tabcolsep}{6pt}
  \begin{threeparttable}
  \begin{tabular}{llll}
    \toprule
    & 3C~\citep{zhang2024spoofdia} & Attractor tokens~\citep{koo2025token} & \cellcolor{oursbg}E2E-SD (ours) \\
    \midrule
    Training data     & PartialSpoof train   & PartialSpoof train   & \cellcolor{oursbg}PartialSpoof train \\
    Train / dev / eval & 25,380 / 24,844 / 71,237 & n/r            & \cellcolor{oursbg}25,380 / 24,844 / 71,237 \\
    Frame resolution  & 20\,ms               & 20\,ms               & \cellcolor{oursbg}20\,ms \\
    Non-speech        & not scored           & not scored           & \cellcolor{oursbg}not scored \\
    SSL front-end     & w2v2-L (317M)        & XLS-R (n/r)          & \cellcolor{oursbg}WavLM-L (315M) \\
    Back-end           & gMLP                 & gMLP                 & \cellcolor{oursbg}gMLP \\
    Loss              & n/r                  & P2SGrad              & \cellcolor{oursbg}scaled CE + BCE \\
    Number of classes & oracle $K$           & oracle $K$           & \cellcolor{oursbg}estimated $\hat K$ \\
    Clustering        & AHC                  & AHC                  & \cellcolor{oursbg}none \\
    Model selection   & n/r                  & best dev             & \cellcolor{oursbg}lowest dev loss \\
    Runs reported     & 1                    & 1                    & \cellcolor{oursbg}3 \\
    $\JER$            & pooled               & pooled               & \cellcolor{oursbg}pooled (Eq.~\ref{eq:jer}) \\
    $\JI$             & all files (Eq.~3)    & all files (Eq.~15)   & \cellcolor{oursbg}files with bonafide \\
    \bottomrule
  \end{tabular}
  \tabnote{n/r: not reported. 3C split sizes and front-end size: from the PartialSpoof paper~\citep{zhang2023partialspoof}. w2v2-L: wav2vec~2.0 Large pre-trained on Libri-Light, CommonVoice, Switchboard and Fisher; XLS-R: checkpoint not stated by \citet{koo2025token}. Non-speech: removed in training except ConP. Pooled: over (utterance, method) pairs. Handling of utterances without bonafide speech is not stated in either paper.}
  \end{threeparttable}
\end{table}

\subsection{Transfer and Detection Sets}
\label{app:transfer_sets}

Table~\ref{tab:transfer_sets} lists the sets outside PartialSpoof.
For LlamaPS, we convert the segment time stamps to the PartialSpoof label format and score the whole signal, because LlamaPS has no voice activity labels.
For diarization, each partially spoofed utterance has two classes (bonafide and one spoofing method).
We use all 76,228 utterances of part a for EER, as \citet{luong2025llamapartialspoof} do for MRM.
\citet{luong2025robust} use the same 76,228 utterances (their Table~I), and \citet{mao2026sal} also use the cross-fade part.

\begin{table}[H]
  \caption{Evaluation sets outside PartialSpoof. None of them is used for training or model selection.}
  \label{tab:transfer_sets}
  \centering\small
  \setlength{\tabcolsep}{4pt}
  \begin{threeparttable}
  \begin{tabular}{@{}>{\raggedright\arraybackslash}p{3.0cm}>{\raggedright\arraybackslash}p{2.6cm}r>{\raggedright\arraybackslash}p{5.6cm}@{}}
    \toprule
    Set & Used for & Utterances & Content \\
    \midrule
    LlamaPS part a (cf) & localization, detection, diarization & 76,228 & 10,573 bonafide, 33,461 fully fake, 32,194 partially fake (cross-fade splices) \\
    LlamaPS part b (cp) & localization, diarization & 32,194 & partially fake, cut-and-paste splices \\
    LlamaPS part b (oa) & localization, diarization & 32,194 & partially fake, overlap-add splices \\
    PartialEdit v1.1 eval & localization, diarization & 53,727 & partially fake, neural speech editing, edited segments spliced into bonafide recordings \\
    ASVspoof 2019 LA eval & detection & 71,237 & bonafide and fully spoofed \\
    ASVspoof 2021 LA eval & detection & 181,566 & as 2019 LA, transmitted over telephony codecs \\
    ASVspoof 2021 DF eval & detection & 611,829 & compressed audio, more vocoders \\
    In-the-Wild & detection & 31,779 & real-world bonafide speech and deepfakes \\
    \bottomrule
  \end{tabular}
  \tabnote{Sources: LlamaPartialSpoof (LlamaPS)~\citep{luong2025llamapartialspoof}; PartialEdit~\citep{zhang2025partialedit}; ASVspoof 2019~\citep{todisco2019asvspoof} and 2021~\citep{yamagishi2021asvspoof}; In-the-Wild~\citep{mueller2022generalize}. LlamaPS has no voice activity labels, so silence is scored. All partially fake LlamaPS and PartialEdit utterances contain one spoofing generator ($K{=}2$). PartialEdit: evaluation split of version 1.1, whose speakers follow the ASVspoof 2019 evaluation split; only the editors E1 (VoiceCraft) and E2 (SSR-Speech) are released.}
  \end{threeparttable}
\end{table}

\section{Ablation Studies}
\label{app:ablations}

Each ablation changes one factor of the default configuration (Table~\ref{tab:hparams}) and is trained with three seeds (777, 1234, 2025).
The default row is the E2E-SD row of Table~\ref{tab:main} and is shaded in every table.
All values are on the PartialSpoof evaluation set with estimated $K$, mean\,$\pm$\,std over three runs.
\best{Bold} marks the best value in a column and \second{underline} the second best.

\subsection{Assignment Loss}
\label{app:abl_loss}

All four losses act on the similarity matrix $\mathbf{S}$ (Eq.~\ref{eq:cos_sim}) and are averaged over the frames $\mathcal{V}$ that have a target.
We write $y_t = \pi^*(c_t)$ for the attractor matched to the class of frame $t$ (Eq.~\ref{eq:hungarian}).

\paragraph{CE on cosine.}
The softmax cross-entropy on the raw similarities:
\begin{equation}
  \Lassign^{\text{CE}} = -\frac{1}{|\mathcal{V}|}\sum_{t\in\mathcal{V}} \log \frac{\exp(S_{t,y_t})}{\sum_{k=0}^{K-1}\exp(S_{t,k})}.
  \label{eq:loss_ce}
\end{equation}
The logits are the raw similarities in $[-1,1]$.
The loss therefore cannot fall below $-\log\big(e/(e+(K{-}1)\,e^{-1})\big)$, its value when the target logit is 1 and the other logits are $-1$; this bound grows with $K$ (0.67 for $K{=}8$).

\paragraph{Scaled CE (default).}
The loss of Eq.~\ref{eq:loss_ce} on the scaled similarities $\hat S_{t,k} = wS_{t,k} + b$ (Eqs.~\ref{eq:scaled_sim} and~\ref{eq:assign}).
The scale removes the bound in practice: the development assignment loss drops from 0.31 to 0.04.

\paragraph{Contrastive (GE2E-style).}
Following the contrast loss of GE2E~\citep{wan2018ge2e}, on the same scaled similarities $\hat S$:
\begin{equation}
  \Lassign^{\text{con}} = \frac{1}{|\mathcal{V}|}\sum_{t\in\mathcal{V}} \Big[\,1 - \sigma\big(\hat S_{t,y_t}\big) + \max_{k\neq y_t} \sigma\big(\hat S_{t,k}\big)\Big],
  \label{eq:loss_con}
\end{equation}
where $\sigma$ is the sigmoid.
Each frame receives gradient only from its target attractor and from the most similar other attractor.

\paragraph{P2SGrad.}
P2SGrad~\citep{zhang2019p2sgrad}, the loss used by \citet{koo2025token}, regresses each similarity to its one-hot target:
\begin{equation}
  \Lassign^{\text{P2S}} = \frac{1}{|\mathcal{V}|\,K}\sum_{t\in\mathcal{V}}\sum_{k=0}^{K-1} \big(S_{t,k} - \mathbf{1}[k = y_t]\big)^2.
  \label{eq:loss_p2s}
\end{equation}
Each attractor is compared with its target independently of the other attractors.

Scaled CE gives the best value in every column of Table~\ref{tab:abl_loss}.
Plain CE is second in every column except $K$-acc.; its $\JER$ is 2.73 points higher.
P2SGrad and the contrastive loss are 5.0--5.8 points worse in $\JER$.
These two losses compare each attractor with the target independently.
They do not use the competition between all attractors that the softmax gives.

\begin{table}[H]
  \caption{Assignment loss (WavLM-L + gMLP, estimated $K$). Mean\,$\pm$\,std over three runs.}
  \label{tab:abl_loss}
  \centering\small
  \setlength{\tabcolsep}{2.5pt}
  \begin{tabular}{lcccccc}
    \toprule
    Loss & $\JI\downarrow$ & $\JER\downarrow$ & \Kn{Known}\,$\downarrow$ & \Va{Varied}\,$\downarrow$ & \Un{Unknown}\,$\downarrow$ & $K$-acc.\,$\uparrow$ \\
    \midrule
    CE on cosine             & \second{\ms{9.30}{0.74}} & \second{\ms{18.74}{1.14}} & \second{\ms{10.43}{0.46}} & \second{\ms{10.19}{0.26}} & \second{\ms{26.36}{1.90}} & \ms{86.83}{1.57} \\
    Contrastive (GE2E-style)~\tcite{wan2018ge2e} & \ms{10.54}{0.29} & \ms{21.01}{0.99} & \ms{13.20}{1.77} & \ms{12.24}{1.40} & \ms{28.60}{1.10} & \ms{84.22}{0.80} \\
    P2SGrad~\tcite{zhang2019p2sgrad} & \ms{10.00}{1.45} & \ms{21.80}{1.83} & \ms{15.26}{0.92} & \ms{13.53}{0.48} & \ms{28.74}{3.10} & \second{\ms{86.99}{1.76}} \\
    \oursrow
    Scaled CE ($wS{+}b$, ours) & \best{\ms{7.77}{0.68}} & \best{\ms{16.01}{0.95}} & \best{\ms{9.55}{0.42}} & \best{\ms{8.70}{0.24}} & \best{\ms{22.33}{1.77}} & \best{\ms{88.44}{1.65}} \\
    \bottomrule
  \end{tabular}
\end{table}

\subsection{Back-end Architecture}
\label{app:abl_backend}

The back-ends have 5 layers and $d{=}512$.
The Transformer uses 8 heads and $d_{\text{ff}}{=}2048$.
The Conformer adds a depthwise convolution with kernel size 31.
gMLP and the Transformer give comparable results (Table~\ref{tab:abl_backend}).
gMLP has the better mean in every column.
The Conformer is 2.05 points worse in $\JER$.
The back-ends do not have the same number of parameters, so this comparison does not separate architecture from capacity.

\begin{table}[H]
  \caption{Sequence back-end (WavLM-L, scaled CE, estimated $K$). Params: back-end parameters used in the forward pass.}
  \label{tab:abl_backend}
  \centering\small
  \setlength{\tabcolsep}{3.2pt}
  \begin{tabular}{lccccccr}
    \toprule
    Back-end & $\JI\downarrow$ & $\JER\downarrow$ & \Kn{Known}\,$\downarrow$ & \Va{Varied}\,$\downarrow$ & \Un{Unknown}\,$\downarrow$ & $K$-acc.\,$\uparrow$ & Params \\
    \midrule
    \oursrow
    gMLP~\tcite{liu2021gmlp} (default) & \best{\ms{7.77}{0.68}} & \best{\ms{16.01}{0.95}} & \best{\ms{9.55}{0.42}} & \best{\ms{8.70}{0.24}} & \best{\ms{22.33}{1.77}} & \best{\ms{88.44}{1.65}} & 24.51M \\
    Transformer~\tcite{vaswani2017attention} & \second{\ms{8.42}{1.15}} & \second{\ms{17.41}{1.26}} & \second{\ms{10.41}{0.69}} & \second{\ms{9.45}{0.90}} & \second{\ms{24.27}{2.41}} & \second{\ms{87.47}{1.44}} & 16.29M \\
    Conformer~\tcite{gulati2020conformer} & \ms{8.78}{0.83} & \ms{18.06}{1.70} & \ms{11.72}{1.28} & \ms{9.90}{1.13} & \ms{24.86}{2.40} & \ms{86.37}{1.75} & 30.83M \\
    \bottomrule
  \end{tabular}
\end{table}

\subsection{SSL Front-End}
\label{app:abl_ssl}

\begin{figure}[H]
  \centering
  \includegraphics[width=\linewidth]{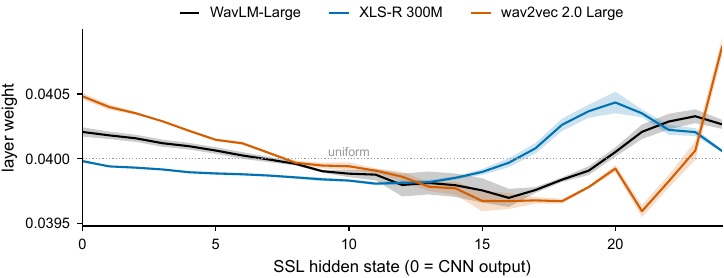}
  \caption{Learned SSL layer weights $\alpha_l$ (mean\,$\pm$\,std over three runs). Dotted: uniform weight $1/25$.}
  \label{fig:layer_weights}
\end{figure}

We compare the front-end of 3C (wav2vec2-Large; the fairseq checkpoint pre-trained on LL+CV+SWBD+Fisher), XLS-R 300M (the family used by \citet{koo2025token}) and WavLM-Large.
All are fine-tuned with scaled CE.
WavLM-Large gives the lowest $\JER$ (Table~\ref{tab:abl_ssl}).
XLS-R gives the lowest Known error.
With a frozen WavLM-Large, $\JER$ increases by 9.0 points and the Known error triples.
The front-end must therefore adapt to the task.
The $K$ estimate does not need this adaptation (88.15\% with the frozen front-end).
The learned layer weights $\alpha_l$ stay close to uniform for all three front-ends (Figure~\ref{fig:layer_weights}).
They are within 1.0\% (WavLM-Large), 1.2\% (XLS-R) and 2.3\% (wav2vec2-Large) of $1/25$.
The front-end adapts through fine-tuning rather than through the choice of layers.

\begin{table}[H]
  \caption{SSL front-end (gMLP, scaled CE, estimated $K$). wav2vec2-Large~\citep{baevski2020wav2vec}: the checkpoint of 3C (LL+CV+SWBD+Fisher); XLS-R~\citep{babu2022xlsr}; WavLM~\citep{chen2022wavlm}.}
  \label{tab:abl_ssl}
  \centering\small
  \setlength{\tabcolsep}{2.8pt}
  \begin{tabular}{lcccccc}
    \toprule
    SSL front-end & $\JI\downarrow$ & $\JER\downarrow$ & \Kn{Known}\,$\downarrow$ & \Va{Varied}\,$\downarrow$ & \Un{Unknown}\,$\downarrow$ & $K$-acc.\,$\uparrow$ \\
    \midrule
    wav2vec2-Large~\tcite{baevski2020wav2vec} & \ms{11.56}{0.39} & \ms{21.90}{0.56} & \ms{12.26}{0.68} & \ms{12.77}{0.66} & \ms{30.25}{0.47} & \ms{84.53}{1.20} \\
    XLS-R 300M~\tcite{babu2022xlsr} & \second{\ms{10.77}{1.74}} & \second{\ms{19.60}{1.60}} & \best{\ms{7.94}{0.19}} & \second{\ms{10.18}{0.70}} & \ms{28.69}{2.75} & \ms{82.27}{2.44} \\
    \oursrow
    WavLM-Large~\tcite{chen2022wavlm} (default) & \best{\ms{7.77}{0.68}} & \best{\ms{16.01}{0.95}} & \second{\ms{9.55}{0.42}} & \best{\ms{8.70}{0.24}} & \best{\ms{22.33}{1.77}} & \best{\ms{88.44}{1.65}} \\
    WavLM-Large, frozen                       & \ms{11.46}{0.07} & \ms{25.01}{0.29} & \ms{28.52}{0.86} & \ms{18.98}{0.50} & \second{\ms{27.69}{0.18}} & \second{\ms{88.15}{0.28}} \\
    \bottomrule
  \end{tabular}
\end{table}

\subsection{Localization Supervision}
\label{app:abl_loc}

E2E-SD has no localization loss: bonafide versus spoof is decided by the assignment to $\mathbf{a}_0$.
We test two additions with $\lambda{=}1.0$.
(i) A separate linear head on $\mathbf{e}_t$ trained with BCE on bonafide versus spoof.
(ii) $\Lbin$: BCE on the $\mathbf{a}_0$ score $w_b\cos(\mathbf{e}_t,\mathbf{a}_0)+b_b$ on every frame of the utterances that contain bonafide, non-speech included, as in the localization labels.
Adding $\Lbin$ reduces seg-EER from 16.02 to 10.32 and utt-EER from 1.31 to 0.96.
However, $\JER$ increases by 2.46 points (Table~\ref{tab:abl_loc}).
The seg-EER does not depend on a threshold, while $\JER$ uses one hard decision per frame (the argmax over attractors).
At that decision point the speech-frame bonafide / spoof error does not improve with $\Lbin$ (7.77\% with vs.\ 6.43\% without).
A lower seg-EER therefore does not give a lower diarization error.

\begin{table}[H]
  \caption{Localization supervision (estimated $K$). seg-EER and utt-EER use the $\mathbf{a}_0$ score in every row.}
  \label{tab:abl_loc}
  \centering\small
  \setlength{\tabcolsep}{3.5pt}
  \begin{tabular}{lcccccc}
    \toprule
    Variant & $\JI\downarrow$ & $\JER\downarrow$ & \Un{Unknown}\,$\downarrow$ & $K$-acc.\,$\uparrow$ & seg-EER\,$\downarrow$ & utt-EER\,$\downarrow$ \\
    \midrule
    \oursrow
    none (default)                   & \best{\ms{7.77}{0.68}} & \best{\ms{16.01}{0.95}} & \best{\ms{22.33}{1.77}} & \best{\ms{88.44}{1.65}} & \ms{16.02}{1.00} & \second{\ms{1.31}{0.15}} \\
    separate linear head (BCE)    & \second{\ms{8.54}{0.93}} & \second{\ms{17.99}{0.94}} & \second{\ms{24.72}{1.50}} & \second{\ms{87.88}{1.65}} & \second{\ms{15.98}{2.74}} & \ms{1.81}{0.02} \\
    $\Lbin$ on $\mathbf{a}_0$              & \ms{9.22}{0.75} & \ms{18.47}{0.73} & \ms{25.99}{1.49} & \ms{85.55}{0.42} & \best{\ms{10.32}{0.90}} & \best{\ms{0.96}{0.12}} \\
    \bottomrule
  \end{tabular}
\end{table}

\subsection{Oracle versus Estimated Number of Classes}
\label{app:abl_oracle}

With the oracle $K$, $\JER$ is 0.08 points lower (Table~\ref{tab:abl_oracle}).
The class count estimate costs less than the run-to-run variation.

\begin{table}[H]
  \caption{Oracle versus estimated number of classes (WavLM-L + gMLP).}
  \label{tab:abl_oracle}
  \centering\small
  \setlength{\tabcolsep}{3.5pt}
  \begin{tabular}{lcccccc}
    \toprule
    $K$ & $\JI\downarrow$ & $\JER\downarrow$ & \Kn{Known}\,$\downarrow$ & \Va{Varied}\,$\downarrow$ & \Un{Unknown}\,$\downarrow$ & $\Delta\JER$ \\
    \midrule
    oracle            & \ms{7.75}{0.69} & \ms{15.93}{0.93} & \ms{9.45}{0.44} & \ms{8.61}{0.19} & \ms{22.25}{1.73} & \na \\
    \oursrow
    estimated (default)       & \ms{7.77}{0.68} & \ms{16.01}{0.95} & \ms{9.55}{0.42} & \ms{8.70}{0.24} & \ms{22.33}{1.77} & \loss{+0.08} \\
    \bottomrule
  \end{tabular}
\end{table}

\subsection{Bonafide Anchoring}
\label{app:abl_anchor}

Without the anchor, the Hungarian matching runs over all $K$ classes, and any attractor can be matched to the bonafide class.
The fixed anchor lowers $\JER$ by 2.48 points and $\JI$ by 2.65 points (Table~\ref{tab:abl_anchor}).
The largest change is on unknown attacks ($-$4.18).
It also makes $\mathbf{a}_0$ the bonafide attractor in 99.5\% of the utterances.
Free matching gives 83.9\%, with large variation between runs.
With free matching, the $\mathbf{a}_0$ score is therefore not a reliable localization score (seg-EER 37.19\,$\pm$\,16.57).
With the anchor, the same score gives seg-EER 16.02\,$\pm$\,1.00 without any localization training.
The utterance EER is slightly lower with free matching (1.06 vs.\ 1.31).

\begin{table}[H]
  \caption{Bonafide anchoring. Free: matching over all $K$ classes ($K!$). Fixed: $\mathbf{a}_0$ is bonafide; only spoof classes are matched ($(K{-}1)!$). $\mathbf{a}_0{=}$bona: utterances whose bonafide speech is assigned to $\mathbf{a}_0$. EERs use the $\mathbf{a}_0$ score. $\Delta$ is computed before rounding.}
  \label{tab:abl_anchor}
  \centering\small
  \setlength{\tabcolsep}{3pt}
  \begin{tabular}{lccccccc}
    \toprule
    Matching & $\JI\downarrow$ & $\JER\downarrow$ & \Un{Unknown}\,$\downarrow$ & $K$-acc.\,$\uparrow$ & $\mathbf{a}_0{=}$bona\,$\uparrow$ & seg-EER\,$\downarrow$ & utt-EER\,$\downarrow$ \\
    \midrule
    free ($K!$)              & \ms{10.41}{1.23} & \ms{18.50}{1.41} & \ms{26.51}{2.44} & \ms{85.75}{2.05} & \ms{83.93}{13.48} & \ms{37.19}{16.57} & \best{\ms{1.06}{0.24}} \\
    \oursrow
    fixed $\mathbf{a}_0$ (default)       & \best{\ms{7.77}{0.68}} & \best{\ms{16.01}{0.95}} & \best{\ms{22.33}{1.77}} & \best{\ms{88.44}{1.65}} & \best{\ms{99.47}{0.10}} & \best{\ms{16.02}{1.00}} & \ms{1.31}{0.15} \\
    \midrule
    $\Delta$ (fixed $-$ free) & \gain{$-$2.65} & \gain{$-$2.48} & \gain{$-$4.18} & \gain{+2.69} & \gain{+15.54} & \gain{$-$21.17} & \loss{+0.25} \\
    \bottomrule
  \end{tabular}
\end{table}

\subsection{Existence Threshold and Attractor Budget at Inference}
\label{app:abl_theta}

The budget $\Kmax$ does not change training (App.~\ref{app:config}).
We therefore vary $\theta$ and $\Kmax$ at inference on the same models, from stored attractor scores (Table~\ref{tab:abl_theta}).
For $\theta$ from 0.3 to 0.7, $\JER$ stays within 15.99--16.03\% and $K$-acc.\ within 88.21--88.64\%.
A lower $\theta$ trades under-estimation for over-estimation (under 10.56 $\to$ 9.82\%, over 1.22 $\to$ 1.54\%).
It also raises the $K{\ge}4$ accuracy (21.3 $\to$ 30.2\%), but this change is smaller than its variation between runs.
Any $\Kmax \in \{6,7,8,11\}$ gives the same results within 0.001 points.
With the default $\theta$, $\hat K$ is at most 6 on PartialSpoof and on all transfer sets for every run.
Even at $\theta{=}0.3$, it never reaches 8.
The estimate of $K$ is therefore not tuned through $\theta$ or $\Kmax$.

\begin{table}[H]
  \caption{Existence threshold $\theta$ at inference (same trained models, stored scores; $\Kmax{=}8$). $\Kmax \in \{6,7,8,11\}$ gives the same values within 0.001 points for every $\theta$. This is because $\hat K$ exceeds 6 in at most 4 utterances (at $\theta{=}0.3$) and never reaches 8. Default: $\theta{=}0.5$ (shaded).}
  \label{tab:abl_theta}
  \centering\small
  \setlength{\tabcolsep}{3pt}
  \begin{tabular}{lccccccc}
    \toprule
    $\theta$ & $\JI\downarrow$ & $\JER\downarrow$ & \Un{Unknown}\,$\downarrow$ & $K$-acc.\,$\uparrow$ & $K{\ge}4$ acc.\,$\uparrow$ & under\,$\downarrow$ & over\,$\downarrow$ \\
    \midrule
    0.3 & \ms{7.77}{0.68} & \ms{15.99}{0.94} & \ms{22.31}{1.76} & \ms{88.64}{1.58} & \ms{30.25}{11.55} & \ms{9.82}{2.02} & \ms{1.54}{0.56} \\
    0.4 & \ms{7.77}{0.68} & \ms{16.00}{0.94} & \ms{22.32}{1.76} & \ms{88.53}{1.62} & \ms{27.70}{11.16} & \ms{10.02}{2.04} & \ms{1.44}{0.52} \\
    \rowcolor{oursbg}
    0.5 & \ms{7.77}{0.68} & \ms{16.01}{0.95} & \ms{22.33}{1.77} & \ms{88.44}{1.65} & \ms{25.60}{10.75} & \ms{10.20}{2.06} & \ms{1.36}{0.50} \\
    0.6 & \ms{7.77}{0.68} & \ms{16.02}{0.95} & \ms{22.33}{1.77} & \ms{88.34}{1.68} & \ms{23.46}{10.47} & \ms{10.37}{2.08} & \ms{1.29}{0.48} \\
    0.7 & \ms{7.77}{0.69} & \ms{16.03}{0.96} & \ms{22.35}{1.78} & \ms{88.21}{1.71} & \ms{21.28}{10.27} & \ms{10.56}{2.10} & \ms{1.22}{0.46} \\
    \bottomrule
  \end{tabular}
\end{table}

\subsection{Existence Training}
\label{app:abl_exist}

Utterances with $K\ge4$ are rare in training (1,001 of 25,380).
We test two common remedies: a class-weighted existence BCE, and balanced sampling that fills 20\% of each epoch with $K\ge4$ utterances.

\paragraph{Weighted BCE.}
Each term of Eq.~\eqref{eq:exist} is multiplied by a weight $\omega_k$ that is larger for attractor indices with few positive targets in the mini-batch:
\begin{equation}
  \begin{gathered}
    \Lexist^{\omega} = -\frac{1}{K{+}1} \sum_{k=0}^{K} \omega_k \big[ y_k \log p_k + (1 - y_k) \log(1 - p_k) \big], \\[4pt]
    \omega_k =
    \begin{cases}
      \min(1/r_k,\, \omega_{\max}) & \text{if } 0 < r_k < \tau, \\
      1 & \text{otherwise},
    \end{cases}
    \qquad
    r_k = \frac{1}{B} \sum_{i=1}^{B} \mathbf{1}[k < K_i],
  \end{gathered}
  \label{eq:wbce}
\end{equation}
where $r_k$ is the share of the $B$ utterances in a mini-batch whose existence target at index $k$ is 1.
We use $\tau{=}0.1$ and $\omega_{\max}{=}50$.

\paragraph{Balanced sampling.}
The training set $\mathcal{D}_{\text{train}}$ is split by the number of classes, and the $K\ge4$ set is oversampled so that it forms a share $\alpha$ of each epoch, $N_{\text{target}}/(N_{\text{low}}+N_{\text{target}}) = \alpha$:
\begin{equation}
  \begin{alignedat}{2}
    \mathcal{S}_{\text{low}} &= \{ i \in \mathcal{D}_{\text{train}} \mid K_i \le 3 \}, &\qquad \mathcal{S}_{\text{high}} &= \{ i \in \mathcal{D}_{\text{train}} \mid K_i \ge 4 \}, \\[2pt]
    N_{\text{target}} &= \Big\lfloor \frac{\alpha}{1-\alpha}\, N_{\text{low}} \Big\rfloor, &\qquad R &= \Big\lceil \frac{N_{\text{target}}}{N_{\text{high}}} \Big\rceil, \\[2pt]
    \mathcal{S}_{\text{aug}} &= (\mathcal{S}_{\text{high}} \times R)[0{:}N_{\text{target}}], &\qquad \mathcal{S}_{\text{epoch}} &= \mathrm{Perm}(\mathcal{S}_{\text{low}} \cup \mathcal{S}_{\text{aug}}),
  \end{alignedat}
  \label{eq:balanced}
\end{equation}
where $N_{\text{low}} = |\mathcal{S}_{\text{low}}|$ and $N_{\text{high}} = |\mathcal{S}_{\text{high}}|$, $\mathcal{S}_{\text{high}} \times R$ repeats $\mathcal{S}_{\text{high}}$ $R$ times in the same order, $[0{:}N_{\text{target}}]$ keeps the first $N_{\text{target}}$ items, and $\mathrm{Perm}$ is a new random permutation in each epoch.
With $\alpha{=}0.2$, $N_{\text{low}}{=}24{,}379$ and $N_{\text{high}}{=}1{,}001$, we get $N_{\text{target}}{=}6{,}094$, $R{=}7$ and 30,473 utterances per epoch.

\paragraph{Results.}
Neither changes $K{\ge}4$ accuracy by more than the variation between runs of the default model (25.60\,$\pm$\,10.75; Table~\ref{tab:abl_exist}).
Balanced sampling gives the smallest variation between runs in $\JER$ and $K$-acc.\ and the lowest under-estimation, at the cost of more over-estimation and 20\% more training utterances per epoch.

\begin{table}[H]
  \caption{Training of the existence head for rare large $K$. $K{\ge}4$: $K$-acc.\ on the 2,332 evaluation utterances with $K\ge4$. Under / over: share of utterances with $\hat K<K$ / $\hat K>K$.}
  \label{tab:abl_exist}
  \centering\small
  \setlength{\tabcolsep}{3.5pt}
  \begin{tabular}{lccccc}
    \toprule
    Existence training & $\JER\downarrow$ & $K$-acc.\,$\uparrow$ & $K{\ge}4$ acc.\,$\uparrow$ & under\,$\downarrow$ & over\,$\downarrow$ \\
    \midrule
    \oursrow
    unweighted BCE (default)                  & \best{\ms{16.01}{0.95}} & \best{\ms{88.44}{1.65}} & \ms{25.60}{10.75} & \second{\ms{10.20}{2.06}} & \best{\ms{1.36}{0.50}} \\
    weighted BCE                           & \ms{17.99}{1.42} & \ms{86.83}{2.02} & \second{\ms{32.00}{3.57}} & \ms{11.51}{2.17} & \second{\ms{1.65}{0.19}} \\
    balanced sampling ($K{\ge}4$ at 20\%)  & \second{\ms{16.55}{0.32}} & \second{\ms{88.03}{0.82}} & \best{\ms{34.62}{14.06}} & \best{\ms{8.81}{2.08}} & \ms{3.16}{2.19} \\
    \bottomrule
  \end{tabular}
\end{table}

\subsection{EDA Time Shuffle}
\label{app:abl_shuffle}

As in EEND-EDA~\citep{horiguchi2020endtoend,horiguchi2022encoder}, the frames fed to the EDA encoder are shuffled during training and inference.
Without the shuffle, $\JER$ increases by 2.49 points (Table~\ref{tab:abl_shuffle}).

\begin{table}[H]
  \caption{Time shuffle of the frames fed to the EDA encoder during training and inference.}
  \label{tab:abl_shuffle}
  \centering\small
  \begin{tabular}{lcccc}
    \toprule
    Time shuffle & $\JI\downarrow$ & $\JER\downarrow$ & \Un{Unknown}\,$\downarrow$ & $K$-acc.\,$\uparrow$ \\
    \midrule
    \oursrow
    on (default) & \best{\ms{7.77}{0.68}} & \best{\ms{16.01}{0.95}} & \best{\ms{22.33}{1.77}} & \best{\ms{88.44}{1.65}} \\
    off       & \ms{9.22}{0.70} & \ms{18.50}{0.40} & \ms{25.88}{1.26} & \ms{86.56}{0.59} \\
    \bottomrule
  \end{tabular}
\end{table}

\subsection{Non-Speech Frames in Training}
\label{app:abl_nonspeech}

By default, non-speech frames other than ConP have no target, following 3C.
If they take the label of the segment that contains them, the non-speech EER drops from 23.26 to 13.90 and seg-EER from 16.02 to 14.90.
However, $\JER$ and utt-EER become worse with this change (Table~\ref{tab:abl_nonspeech}).

\begin{table}[H]
  \caption{Non-speech frames in the assignment loss. Ignored: non-speech frames (except ConP) have no target. Label of the segment: they take the label of the segment that contains them. Non-speech EER: 20\,ms EER on non-speech frames only.}
  \label{tab:abl_nonspeech}
  \centering\small
  \setlength{\tabcolsep}{3.5pt}
  \begin{tabular}{lccccc}
    \toprule
    Non-speech frames & $\JI\downarrow$ & $\JER\downarrow$ & seg-EER\,$\downarrow$ & non-speech EER\,$\downarrow$ & utt-EER\,$\downarrow$ \\
    \midrule
    \oursrow
    ignored (default)       & \best{\ms{7.77}{0.68}} & \best{\ms{16.01}{0.95}} & \ms{16.02}{1.00} & \ms{23.26}{1.16} & \best{\ms{1.31}{0.15}} \\
    label of the segment & \ms{8.91}{0.61} & \ms{17.69}{0.99} & \best{\ms{14.90}{2.19}} & \best{\ms{13.90}{3.10}} & \ms{1.76}{0.20} \\
    \bottomrule
  \end{tabular}
\end{table}

\section{Additional Results}
\label{app:results}

\subsection{Per-Method $\JER$}
\label{app:permethod}

Table~\ref{tab:permethod} gives $\JER$ for each spoofing method.
E2E-SD has the lowest $\JER$ on all 13 methods.
On A16 the margin is small (10.65\,$\pm$\,0.60 vs.\ 11.18).
The three attacks with the highest 3C error remain the hardest for E2E-SD: A10 (57.19 $\to$ 42.39), A12 (53.92 $\to$ 33.91) and A15 (59.76 $\to$ 41.34).
For E2E-SD, all other methods are below 15.
The largest absolute reductions are on A11 and A12.
In 3C, adding the localization branch improves $\JI$ but raises the unknown-attack $\JER$ (36.78 $\to$ 49.02).

\begin{table}[H]
  \caption{$\JER$ (\%) per spoofing method on the PartialSpoof evaluation set, ordered by attack group (\Kn{Known}, \Va{Varied}, \Un{Unknown}). 3C uses oracle $K$; E2E-SD estimates $K$. E2E-SD: mean over three runs, std in the row below. \best{Bold}: best per column.}
  \label{tab:permethod}
  \centering
  \footnotesize
  \setlength{\tabcolsep}{2.6pt}
  \begin{threeparttable}
  \begin{tabular}{lccccccccccccc}
    \toprule
    & \multicolumn{2}{c}{\Kn{Known}} & \multicolumn{4}{c}{\Va{Varied}} & \multicolumn{7}{c}{\Un{Unknown}} \\
    \cmidrule(lr){2-3}\cmidrule(lr){4-7}\cmidrule(l){8-14}
    System & \Kn{A16} & \Kn{A19} & \Va{A07} & \Va{A08} & \Va{A09} & \Va{A17} & \Un{A10} & \Un{A11} & \Un{A12} & \Un{A13} & \Un{A14} & \Un{A15} & \Un{A18} \\
    \midrule
    3C (Mul)~\tcite{zhang2024spoofdia} & 11.18 & 11.99 & 13.99 & 11.02 & 16.90 & 21.10 & 57.19 & 32.95 & 53.92 & 16.61 & 14.81 & 59.76 & 18.59 \\
    3C (Mul+Bin)~\tcite{zhang2024spoofdia} & 11.34 & 12.72 & 12.82 & 11.11 & 16.29 & 23.18 & 84.51 & 44.73 & 70.87 & 16.37 & 16.09 & 86.59 & 16.93 \\
    \oursrow
    E2E-SD (ours)       & \best{10.65} & \best{8.27} & \best{7.69} & \best{7.20} & \best{7.30} & \best{13.58} & \best{42.39} & \best{13.27} & \best{33.91} & \best{7.74} & \best{7.52} & \best{41.34} & \best{7.04} \\
    \oursrow
    {\color{mutedgray}\quad $\pm$ std} & {\color{mutedgray}0.60} & {\color{mutedgray}0.28} & {\color{mutedgray}0.20} & {\color{mutedgray}0.18} & {\color{mutedgray}0.75} & {\color{mutedgray}0.22} & {\color{mutedgray}3.53} & {\color{mutedgray}0.99} & {\color{mutedgray}4.40} & {\color{mutedgray}0.74} & {\color{mutedgray}0.87} & {\color{mutedgray}2.98} & {\color{mutedgray}0.32} \\
    \bottomrule
  \end{tabular}
  \tabnote{3C rows: \citet{zhang2024spoofdia}, arXiv Appendix Table~2 (complete rows only). 3C (Mul): Mul-CM as the diarization model, no localization branch (best 3C $\JER$); 3C (Mul+Bin): with a Bin-CM localization branch (the best 3C model).}
  \end{threeparttable}
\end{table}

\subsection{Decomposition of $\JER$}
\label{app:jer_decomp}

Table~\ref{tab:jer_decomp} splits $\JER$ by the kind of error.
Two thirds of $\JER$ (65.0\%) come from localization errors, mostly spoof time that the model labels as bonafide.
Attribution errors between spoofing methods are about one third.

\begin{table}[H]
  \caption{Decomposition of $\JER$ into localization and attribution errors (estimated $K$). Localization: spoof time labeled with the bonafide cluster (spoof $\to$ bona) or bonafide time labeled with a spoof cluster (bona $\to$ spoof). Attribution: spoof time given to another spoof cluster. With uncovered speech (0.04), the parts add up to $\JER$ (\%; mean\,$\pm$\,std over three runs).}
  \label{tab:jer_decomp}
  \centering\small
  \setlength{\tabcolsep}{3.5pt}
  \begin{threeparttable}
  \begin{tabular}{lccccc}
    \toprule
    & & \multicolumn{3}{c}{Localization} & \\
    \cmidrule(lr){3-5}
    Matching & $\JER$ & total & spoof $\to$ bona & bona $\to$ spoof & Attribution \\
    \midrule
    \oursrow
    fixed $\mathbf{a}_0$ (default) & \ms{16.01}{0.95} & \ms{10.41}{0.99} {\scriptsize(65.0\%)} & \ms{8.35}{0.89} & \ms{2.06}{0.19} & \ms{5.56}{0.05} \\
    free               & \ms{18.50}{1.41} & \ms{13.12}{1.74} {\scriptsize(70.9\%)} & \ms{9.98}{1.20} & \ms{3.14}{0.60} & \ms{5.34}{0.33} \\
    \bottomrule
  \end{tabular}
  \tabnote{Shares in parentheses: localization error over $\JER$ (ratio of means). Uncovered speech contributes 0.04 to $\JER$ and is not shown.}
  \end{threeparttable}
\end{table}

\subsection{Outcome of Each Attack Instance}
\label{app:instances}

An instance is one spoofing method in one utterance.
Table~\ref{tab:instances} shows what happens to each instance.
In utterances with $\hat K = K - 1$, 33.1\% of the spoofing-method instances are missed (absorbed by the bonafide cluster).
In these utterances, 9.6\% of the instances are merged with another spoofing method.
Over all instances, a method is missed about three times as often as it is attributed to a wrong cluster (8.3\% vs.\ 3.1\%).
For forensic use, the main risk of E2E-SD is a missed manipulation rather than a wrong attribution.

\begin{table}[H]
  \caption{Outcome of the spoofing-method instances of the evaluation set (mean number per run), by the class-count error of the utterance (\% of row; mean\,$\pm$\,std over three runs). Correct: own cluster. Missed: mostly bonafide cluster. Merged: mostly another spoof cluster.}
  \label{tab:instances}
  \centering\small
  \begin{threeparttable}
  \begin{tabular}{lrccc}
    \toprule
    Utterances & Instances & Correct & Missed $\to$ bonafide & Merged $\to$ other spoof \\
    \midrule
    $\hat K = K$         & 42,444 & \ms{96.4}{0.7} & \ms{3.2}{0.6}  & \ms{0.4}{0.2} \\
    $\hat K = K-1$       & 8,281  & \ms{57.2}{1.4} & \ms{33.1}{1.8} & \ms{9.6}{2.5} \\
    $\hat K \le K-2$     & 2,325  & \ms{55.3}{1.2} & \ms{14.3}{4.0} & \ms{30.4}{3.9} \\
    $\hat K > K$         & 465    & \ms{98.4}{0.8} & \ms{0.5}{0.2}  & \ms{0.6}{0.6} \\
    \midrule
    all                  & 53,515 & \ms{88.6}{0.9} & \ms{8.3}{1.0}  & \ms{3.1}{0.2} \\
    \bottomrule
  \end{tabular}
  \tabnote{For $\hat K > K$, the remaining 0.5\% of the instances have no mapped cluster.}
  \end{threeparttable}
\end{table}

\subsection{Frame--Attractor Similarity in One Utterance}
\label{app:anatomy}

\begin{figure}[H]
  \centering
  \includegraphics[width=\linewidth]{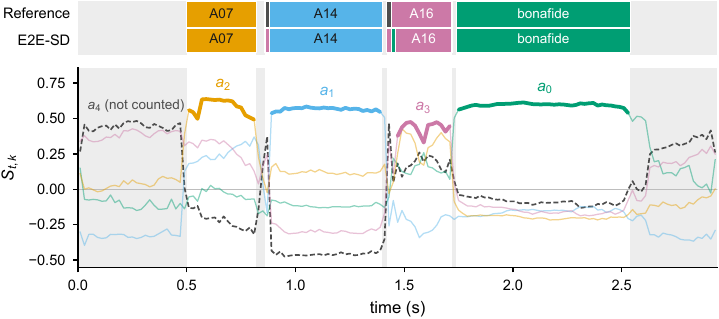}
  \caption{One evaluation utterance with three spoofing methods (A07 varied, A14 unknown, A16 known; one run). Top: reference and E2E-SD output. Bottom: $S_{t,k}$ of the counted attractors, thick where $\mathbf{a}_k$ wins the argmax on a scored frame. Dashed: the first attractor that is not counted ($p_4 < \theta$). Grey: non-speech, not scored.}
  \label{fig:anatomy}
\end{figure}

In the utterance of Figure~\ref{fig:anatomy}, on each scored segment the attractor of its class has the highest similarity.
The other attractors stay well below it.
The model estimates $\hat K{=}4$, and the argmax gives the reference segments.
The only errors are short spans near the segment boundaries.
The fifth attractor is not counted.
Its similarity is high only in non-speech, which is not scored.

\subsection{Qualitative Failure Cases}
\label{app:error_cases}

\begin{figure}[H]
  \centering
  \includegraphics[width=\linewidth]{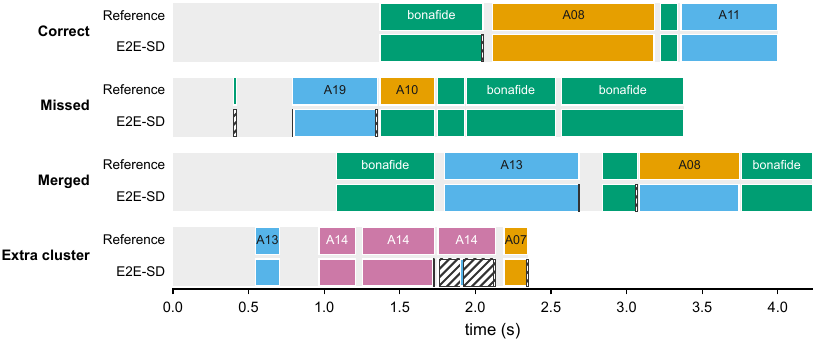}
  \caption{Reference and E2E-SD output for four evaluation utterances with $K{=}4$ ($K$ includes ConP, which is not scored or shown). Correct: $\hat K{=}4$. Missed: the shortest spoofed class is absorbed by bonafide ($\hat K{=}3$). Merged: two spoofed classes share one cluster ($\hat K{=}3$). Extra cluster: one spoofed class is split into two clusters ($\hat K{=}5$). Colours: output clusters mapped to the reference classes; hatched: unmapped cluster. Grey: not scored.}
  \label{fig:error_cases}
\end{figure}

Figure~\ref{fig:error_cases} shows four evaluation utterances with $K{=}4$: a correct case, a missed method, two merged methods and a method split into two clusters.

\section{Cross-Dataset Transfer Details}
\label{app:transfer}

All models in this section are trained on PartialSpoof only (Section~\ref{sec:transfer}).
Every partially spoofed utterance in LlamaPS and PartialEdit contains a single spoofing method, so the correct class count is $K{=}2$.

\subsection{LlamaPS by Splice Method}
\label{app:splice}

LlamaPS joins bonafide and generated speech in three ways: cross-fade (cf, part a), cut-and-paste (cp, part b) and overlap-add (oa, part b).
Table~\ref{tab:splice} gives the results for each method.
The models trained with ConP almost never predict $\hat K{=}2$ ($K$-acc.\ below 5\%) for any of the three splice methods.
The model without ConP predicts it in 91--92\% of the utterances.
The over-estimation therefore does not depend on the splice method.
Without ConP, seg-EER is higher than with ConP for all three splice methods (cf 39.50 vs.\ 37.50, cp 40.76 vs.\ 38.13, oa 39.00 vs.\ 37.64).
In training, ConP is the only class that gives a target to frames between speech segments.
Without it, the model does not learn that these regions carry spoofing information.
Earlier work found such information in the silence of spoofed audio~\citep{zhang2021silence} and showed that models can exploit it as a shortcut~\citep{muller2021silence}.

\begin{table}[H]
  \caption{LlamaPS by splice method (32,194 partially spoofed utterances each, $K{=}2$; \%). seg-EER here uses the partially spoofed utterances only, unlike Table~\ref{tab:transfer} (all 76,228 utterances). Mean\,$\pm$\,std over three runs. \best{Bold}: best; \second{underline}: second best.}
  \label{tab:splice}
  \centering\small
  \begin{threeparttable}
  \begin{tabular}{lccc}
    \toprule
    Model & cross-fade (part a) & cut-and-paste (part b) & overlap-add (part b) \\
    \midrule
    \grouprow{4}{seg-EER\,$\downarrow$}
    \oursrow
    E2E-SD (default) & \best{\ms{37.50}{2.16}} & \best{\ms{38.13}{2.53}} & \best{\ms{37.64}{1.30}} \\
    E2E-SD without ConP & \ms{39.50}{1.42} & \ms{40.76}{2.27} & \ms{39.00}{1.32} \\
    E2E-SD + $\Lbin$ & \second{\ms{38.25}{0.37}} & \second{\ms{38.17}{0.53}} & \second{\ms{38.16}{0.34}} \\
    E2E-SD + non-speech targets & \ms{38.34}{0.76} & \ms{38.90}{1.09} & \ms{38.33}{0.90} \\
    \grouprow{4}{$\JER\downarrow$}
    \oursrow
    E2E-SD (default) & \ms{35.02}{2.10} & \ms{31.72}{2.05} & \ms{36.57}{1.55} \\
    E2E-SD without ConP & \best{\ms{30.34}{0.83}} & \best{\ms{27.74}{1.50}} & \best{\ms{32.13}{0.50}} \\
    E2E-SD + $\Lbin$ & \ms{35.23}{1.29} & \ms{31.07}{1.17} & \ms{37.12}{0.91} \\
    E2E-SD + non-speech targets & \second{\ms{32.17}{0.32}} & \second{\ms{28.14}{0.71}} & \second{\ms{34.17}{0.47}} \\
    \grouprow{4}{$K$-acc.\,$\uparrow$}
    \oursrow
    E2E-SD (default) & \ms{3.15}{1.55} & \ms{2.93}{1.43} & \ms{3.01}{1.65} \\
    E2E-SD without ConP & \best{\ms{91.77}{2.21}} & \best{\ms{91.62}{2.43}} & \best{\ms{91.20}{3.21}} \\
    E2E-SD + $\Lbin$ & \ms{2.03}{1.32} & \ms{2.08}{1.16} & \ms{1.84}{1.34} \\
    E2E-SD + non-speech targets & \second{\ms{3.99}{2.26}} & \second{\ms{4.21}{1.53}} & \second{\ms{3.20}{2.40}} \\
    \bottomrule
  \end{tabular}
  \tabnote{Each row changes one factor of E2E-SD: without ConP (Section~\ref{sec:conp}); $+\,\Lbin$ (App.~\ref{app:abl_loc}); + non-speech targets: non-speech frames take the label of their segment (App.~\ref{app:abl_nonspeech}).}
  \end{threeparttable}
\end{table}

\subsection{PartialEdit}
\label{app:partialedit}

PartialEdit replaces words by neural speech editing and splices the edited segments back into the original audio~\citep{zhang2025partialedit}.
We use the evaluation split of version 1.1: 53,727 utterances from the two released editors, VoiceCraft (E1) and SSR-Speech (E2).
Table~\ref{tab:partialedit} shows the same pattern as LlamaPS for seg-EER and $K$-acc.
ConP lowers seg-EER (32.42 vs.\ 39.78\%) and $\JER$ (61.05 vs.\ 70.33\%).
Without ConP, the model counts the classes correctly in 70.0\% of the utterances (23.1\% with ConP) and has the best $\JI$.
All models have a high $\JER$ (60--75\%), because the edits are short (0.71\,s on average) and the editors are not in the training data.

\begin{table}[H]
  \caption{PartialEdit v1.1 evaluation split (53,727 partially spoofed utterances from the editors E1 and E2, $K{=}2$; \%). Mean\,$\pm$\,std over three runs. \best{Bold}: best; \second{underline}: second best.}
  \label{tab:partialedit}
  \centering\small
  \begin{threeparttable}
  \begin{tabular}{lcccc}
    \toprule
    Model & seg-EER\,$\downarrow$ & $\JER\downarrow$ & $\JI\downarrow$ & $K$-acc.\,$\uparrow$ \\
    \midrule
    \oursrow
    E2E-SD (default) & \best{\ms{32.42}{1.98}} & \best{\ms{61.05}{3.01}} & \ms{48.24}{1.64} & \ms{23.10}{17.23} \\
    E2E-SD without ConP & \second{\ms{39.78}{1.33}} & \ms{70.33}{2.46} & \best{\ms{30.14}{4.42}} & \best{\ms{69.99}{4.24}} \\
    E2E-SD + $\Lbin$ & \ms{44.26}{1.67} & \second{\ms{64.33}{3.52}} & \ms{48.43}{1.00} & \ms{13.93}{6.28} \\
    E2E-SD + non-speech targets & \ms{44.46}{2.87} & \ms{75.46}{2.46} & \second{\ms{40.27}{2.54}} & \second{\ms{24.63}{12.66}} \\
    \bottomrule
  \end{tabular}
  \tabnote{Each row changes one factor of E2E-SD (see Table~\ref{tab:splice}).}
  \end{threeparttable}
\end{table}

\subsection{LlamaPS by Generator}
\label{app:generator}

LlamaPS uses six TTS systems.
Each partially spoofed utterance uses one of them (Table~\ref{tab:generator}).
Localization depends strongly on the generator: seg-EER is 21.55\% for GPT-SoVITS and 72.14\% for YourTTS.
For YourTTS, the seg-EER exceeds 50\% for both models in Table~\ref{tab:generator}.
That is, the $\mathbf{a}_0$ score assigns higher bonafide scores to the spoofed YourTTS frames than to the bonafide frames.
Utterance detection follows a different order: YourTTS is detected well (13.96\%), and JETS is the hardest (37.91\%).
Without ConP, $\JER$ is lower and $K$-acc.\ is 87--94\% for every generator.
The effect of ConP (Section~\ref{sec:conp}) therefore does not come from one generator.

\begin{table}[H]
  \caption{LlamaPS part a by spoofing generator (cross-fade splices; \%). seg-EER, $\JER$ and $K$-acc.\ use the partially spoofed utterances. utt-EER: bonafide (10,573) against all fake utterances of the generator. Mean over three runs. The cp and oa splices give the same pattern.}
  \label{tab:generator}
  \centering\small
  \setlength{\tabcolsep}{3pt}
  \begin{tabular}{@{}lrcccccccc@{}}
    \toprule
    & & \multicolumn{3}{c}{E2E-SD (default)} & \multicolumn{3}{c}{without ConP} & \multicolumn{2}{c}{utt-EER\,$\downarrow$} \\
    \cmidrule(lr){3-5}\cmidrule(lr){6-8}\cmidrule(l){9-10}
    Generator & \#Utt. & seg\,$\downarrow$ & $\JER\downarrow$ & $K$-acc.\,$\uparrow$ & seg\,$\downarrow$ & $\JER\downarrow$ & $K$-acc.\,$\uparrow$ & default & w/o ConP \\
    \midrule
    CosyVoice & 5,262 & 32.20 & 41.60 & 5.24 & 37.34 & 38.59 & 86.56 & 23.47 & 30.42 \\
    ElevenLabs & 5,387 & 40.66 & 42.23 & 4.04 & 43.97 & 37.19 & 89.60 & 23.10 & 29.98 \\
    GPT-SoVITS & 5,387 & 21.55 & 32.67 & 2.39 & 23.32 & 30.93 & 94.03 & 11.25 & 17.40 \\
    JETS (LJSpeech) & 5,387 & 31.87 & 22.01 & 2.56 & 30.75 & 17.29 & 94.16 & 37.91 & 29.69 \\
    XTTS v2 & 5,384 & 28.40 & 34.06 & 2.36 & 32.77 & 30.74 & 92.10 & 12.05 & 19.43 \\
    YourTTS & 5,387 & 72.14 & 37.72 & 2.36 & 70.28 & 27.51 & 94.02 & 13.96 & 19.41 \\
    \midrule
    \textit{all} & 32,194 & 37.50 & 35.02 & 3.15 & 39.50 & 30.34 & 91.77 & 19.71 & 25.10 \\
    \bottomrule
  \end{tabular}
\end{table}

\subsection{Detection by Condition}
\label{app:det_subsets}

Table~\ref{tab:det_subsets} splits the detection EER by condition.
On ASVspoof 2021 LA, the error comes from the transmission channel: 3.97\% without a codec and 41.85\% for GSM.
PartialSpoof contains no codec or channel effects, and we use no augmentation.
On ASVspoof 2021 DF, MP3 compression gives the highest error (21--27\%).
On ASVspoof 2019 LA, the error is concentrated on A10 and varies strongly between runs (5.64 / 26.11 / 31.04\%).
In-the-Wild has no condition labels.

\begin{table}[H]
  \caption{Detection EER (\%) of E2E-SD by condition. 21LA/21DF rows: bonafide and spoof trials of that condition; 19LA rows: all bonafide trials against one attack. Mean\,$\pm$\,std over three runs.}
  \label{tab:det_subsets}
  \centering\small
  \setlength{\tabcolsep}{4pt}
  \begin{minipage}[t]{0.30\linewidth}\centering
    \begin{tabular}[t]{lr}
      \toprule
      21LA channel & EER\,$\downarrow$ \\
      \midrule
      none   & \ms{3.97}{1.24} \\
      a-law  & \ms{14.64}{5.08} \\
      PSTN   & \ms{32.45}{13.78} \\
      G.722  & \ms{5.74}{2.02} \\
      $\mu$-law & \ms{15.04}{4.98} \\
      GSM    & \ms{41.85}{10.22} \\
      Opus   & \ms{26.18}{7.41} \\
      \midrule
      all    & \ms{25.29}{5.73} \\
      eval.\ phase & \ms{25.34}{6.28} \\
      \bottomrule
    \end{tabular}
  \end{minipage}\hfill
  \begin{minipage}[t]{0.33\linewidth}\centering
    \begin{tabular}[t]{lr}
      \toprule
      21DF codec & EER\,$\downarrow$ \\
      \midrule
      none          & \ms{12.18}{2.36} \\
      MP3, low      & \ms{26.87}{7.98} \\
      MP3, high     & \ms{21.37}{6.77} \\
      M4A, low      & \ms{14.58}{3.20} \\
      M4A, high     & \ms{13.72}{2.89} \\
      OGG, low      & \ms{14.36}{3.06} \\
      OGG, high     & \ms{12.40}{2.48} \\
      MP3 + M4A     & \ms{25.98}{7.49} \\
      OGG + M4A     & \ms{14.69}{3.16} \\
      \midrule
      all           & \ms{18.52}{3.98} \\
      eval.\ phase  & \ms{19.58}{4.72} \\
      \bottomrule
    \end{tabular}
  \end{minipage}\hfill
  \begin{minipage}[t]{0.30\linewidth}\centering
    \begin{tabular}[t]{lr}
      \toprule
      19LA attack & EER\,$\downarrow$ \\
      \midrule
      \Va{A07} & \ms{1.54}{1.45} \\
      \Va{A08} & \ms{0.59}{0.14} \\
      \Va{A09} & \ms{0.90}{0.08} \\
      \Un{A10} & \ms{20.93}{13.47} \\
      \Un{A11} & \ms{3.86}{2.18} \\
      \Un{A12} & \ms{0.46}{0.31} \\
      \Un{A13} & \ms{1.87}{1.98} \\
      \Un{A14} & \ms{0.63}{0.19} \\
      \Un{A15} & \ms{1.17}{0.46} \\
      \Kn{A16} & \ms{0.33}{0.19} \\
      \Va{A17} & \ms{0.32}{0.11} \\
      \Un{A18} & \ms{0.34}{0.09} \\
      \Kn{A19} & \ms{0.30}{0.15} \\
      \midrule
      all      & \ms{2.95}{1.23} \\
      \bottomrule
    \end{tabular}
  \end{minipage}
  \par\vspace{4pt}
  \begin{minipage}{\linewidth}\footnotesize eval.\ phase: official evaluation-phase trials only (21LA 148,176; 21DF 533,928). 21DF by source corpus: ASVspoof \ms{17.04}{3.19}, VCC 2018 \ms{29.49}{7.45}, VCC 2020 \ms{19.51}{4.14}. 19LA attack colors: \Kn{known}, \Va{varied}, \Un{unknown} (App.~\ref{app:data}). A10 per run: 5.64 / 26.11 / 31.04.
  \end{minipage}
\end{table}

\section{Per-Run Results}
\label{app:per_run}

Runs with the same configuration and seed are not identical, because cuDNN LSTM and scatter kernels are not deterministic.
Table~\ref{tab:seeds_ablation} lists every run of the ablation and control configurations.

\begin{table}[H]
  \caption{Ablation and control configurations: every run (seeds 777, 1234, 2025; PartialSpoof eval, \%). EERs use the $\mathbf{a}_0$ score.}
  \label{tab:seeds_ablation}
  \centering\footnotesize
  \setlength{\tabcolsep}{3pt}
  \begin{tabular}{@{}llcccccccc@{}}
    \toprule
    Configuration & Seed & $\JI$ & $\JER$ & \Kn{Known} & \Va{Varied} & \Un{Unknown} & $K$-acc. & seg-EER & utt-EER \\
    \midrule
    \multirow{3}{*}{CE} & 777 & 8.82 & 17.89 & 10.17 & 9.89 & 24.98 & 88.41 & 16.46 & 1.22 \\
     & 1234 & 8.93 & 18.31 & 10.16 & 10.26 & 25.56 & 86.80 & 16.78 & 0.84 \\
     & 2025 & 10.15 & 20.04 & 10.96 & 10.40 & 28.53 & 85.28 & 19.54 & 0.97 \\
    \midrule
    \multirow{3}{*}{Contrastive (GE2E)} & 777 & 10.24 & 20.72 & 13.54 & 12.97 & 27.51 & 84.97 & 19.51 & 1.40 \\
     & 1234 & 10.53 & 20.19 & 11.28 & 10.62 & 28.60 & 84.30 & 19.65 & 1.35 \\
     & 2025 & 10.83 & 22.11 & 14.78 & 13.13 & 29.71 & 83.38 & 20.76 & 1.29 \\
    \midrule
    \multirow{3}{*}{P2SGrad} & 777 & 8.42 & 19.76 & 14.93 & 13.02 & 25.27 & 88.78 & 15.48 & 2.05 \\
     & 1234 & 11.26 & 23.30 & 16.30 & 13.59 & 31.25 & 85.26 & 18.73 & 1.01 \\
     & 2025 & 10.33 & 22.34 & 14.55 & 13.97 & 29.69 & 86.94 & 18.02 & 1.05 \\
    \midrule
    \multirow{3}{*}{Transformer back-end} & 777 & 7.09 & 16.10 & 10.79 & 9.72 & 21.52 & 89.10 & 12.48 & 1.14 \\
     & 1234 & 9.08 & 18.61 & 10.81 & 10.19 & 25.98 & 86.93 & 15.26 & 1.26 \\
     & 2025 & 9.09 & 17.51 & 9.61 & 8.44 & 25.32 & 86.38 & 16.51 & 1.22 \\
    \midrule
    \multirow{3}{*}{Conformer back-end} & 777 & 9.72 & 19.97 & 12.72 & 10.82 & 27.63 & 84.43 & 16.96 & 1.33 \\
     & 1234 & 8.17 & 17.48 & 12.17 & 10.24 & 23.43 & 86.87 & 14.24 & 1.05 \\
     & 2025 & 8.44 & 16.72 & 10.28 & 8.65 & 23.51 & 87.82 & 15.27 & 1.21 \\
    \midrule
    \multirow{3}{*}{wav2vec2-Large} & 777 & 11.63 & 22.20 & 12.46 & 13.06 & 30.58 & 85.09 & 19.40 & 1.59 \\
     & 1234 & 11.91 & 22.25 & 12.82 & 13.24 & 30.45 & 83.15 & 19.63 & 1.52 \\
     & 2025 & 11.14 & 21.26 & 11.50 & 12.02 & 29.71 & 85.35 & 18.65 & 1.39 \\
    \midrule
    \multirow{3}{*}{XLS-R 300M} & 777 & 9.17 & 18.15 & 8.04 & 10.09 & 25.97 & 84.38 & 16.45 & 1.44 \\
     & 1234 & 12.63 & 21.32 & 8.08 & 10.92 & 31.47 & 79.60 & 22.41 & 2.15 \\
     & 2025 & 10.52 & 19.32 & 7.72 & 9.53 & 28.63 & 82.82 & 18.55 & 1.26 \\
    \midrule
    \multirow{3}{*}{WavLM-Large frozen} & 777 & 11.53 & 25.31 & 29.44 & 19.30 & 27.82 & 87.84 & 23.33 & 5.48 \\
     & 1234 & 11.47 & 24.96 & 28.38 & 19.23 & 27.48 & 88.20 & 23.61 & 12.96 \\
     & 2025 & 11.39 & 24.75 & 27.74 & 18.41 & 27.77 & 88.41 & 24.14 & 16.80 \\
    \midrule
    \multirow{3}{*}{separate linear loc head} & 777 & 8.65 & 18.19 & 11.08 & 10.27 & 25.06 & 86.54 & 17.30 & 1.84 \\
     & 1234 & 9.40 & 18.82 & 11.00 & 10.71 & 26.01 & 87.37 & 17.81 & 1.79 \\
     & 2025 & 7.55 & 16.96 & 10.43 & 10.02 & 23.07 & 89.72 & 12.82 & 1.81 \\
    \midrule
    \multirow{3}{*}{$\Lbin$ on $\mathbf{a}_0$} & 777 & 8.97 & 18.17 & 9.51 & 10.09 & 25.59 & 85.99 & 9.92 & 0.83 \\
     & 1234 & 10.07 & 19.31 & 9.77 & 10.14 & 27.65 & 85.52 & 11.35 & 1.01 \\
     & 2025 & 8.63 & 17.94 & 9.30 & 10.86 & 24.75 & 85.15 & 9.69 & 1.05 \\
    \midrule
    \multirow{3}{*}{free matching} & 777 & 9.00 & 17.01 & 9.99 & 8.72 & 24.09 & 87.61 & 18.33 & 1.24 \\
     & 1234 & 11.16 & 18.66 & 10.81 & 9.60 & 26.45 & 86.08 & 43.88 & 1.14 \\
     & 2025 & 11.08 & 19.81 & 10.02 & 9.41 & 28.98 & 83.56 & 49.36 & 0.79 \\
    \midrule
    \multirow{3}{*}{weighted BCE} & 777 & 7.88 & 16.81 & 10.14 & 9.59 & 23.14 & 88.93 & 16.73 & 1.55 \\
     & 1234 & 11.22 & 19.57 & 10.36 & 9.65 & 28.26 & 84.90 & 19.70 & 1.20 \\
     & 2025 & 8.94 & 17.59 & 10.05 & 8.73 & 25.17 & 86.67 & 17.10 & 1.20 \\
    \midrule
    \multirow{3}{*}{balanced sampling} & 777 & 7.64 & 16.35 & 11.55 & 9.64 & 21.82 & 87.57 & 16.71 & 2.16 \\
     & 1234 & 8.65 & 16.92 & 11.77 & 9.26 & 23.08 & 87.55 & 16.80 & 1.06 \\
     & 2025 & 7.50 & 16.37 & 11.95 & 9.79 & 21.67 & 88.98 & 17.11 & 2.31 \\
    \midrule
    \multirow{3}{*}{no time shuffle} & 777 & 8.60 & 18.04 & 11.25 & 10.49 & 24.60 & 87.22 & 14.83 & 1.28 \\
     & 1234 & 9.98 & 18.70 & 10.61 & 8.70 & 27.12 & 86.09 & 17.86 & 1.37 \\
     & 2025 & 9.09 & 18.77 & 11.17 & 10.62 & 25.93 & 86.37 & 15.28 & 1.50 \\
    \midrule
    \multirow{3}{*}{non-speech targets} & 777 & 8.33 & 16.90 & 9.54 & 10.16 & 23.13 & 86.46 & 13.49 & 1.93 \\
     & 1234 & 8.85 & 17.36 & 9.50 & 10.15 & 24.02 & 86.48 & 13.78 & 1.54 \\
     & 2025 & 9.54 & 18.81 & 10.00 & 10.46 & 26.43 & 85.16 & 17.42 & 1.82 \\
    \midrule
    \multirow{3}{*}{without ConP} & 777 & 7.07 & 16.12 & 10.47 & 9.47 & 21.80 & 89.64 & 16.08 & 18.68 \\
     & 1234 & 9.87 & 20.93 & 13.74 & 11.82 & 28.55 & 85.34 & 22.98 & 30.25 \\
     & 2025 & 8.53 & 18.09 & 10.94 & 10.08 & 25.04 & 88.27 & 18.54 & 15.01 \\
    \bottomrule
  \end{tabular}
\end{table}

\section{Limitations and Future Directions}
\label{app:limitations}

\paragraph{One training corpus.}
All models are trained on PartialSpoof, to our knowledge the only public corpus with segment-level labels for several spoofing methods per utterance.
Every other partially spoofed corpus we surveyed, including LlamaPartialSpoof and PartialEdit, contains a single spoofing method per utterance.
The budget $\Kmax$ follows the largest class count of this corpus and must be set again for another corpus (App.~\ref{app:config}).

\paragraph{Transfer tests have $K{=}2$.}
Every partially spoofed utterance in LlamaPS and PartialEdit contains a single spoofing method.
The transfer results therefore test localization and the class count, but not the attribution of several methods in one utterance outside PartialSpoof.
This limit comes from the available data, and it leaves an open direction: a benchmark that mixes several spoofing methods in one utterance outside PartialSpoof would test attribution directly.

\paragraph{One speaker per utterance.}
PartialSpoof has one speaker per utterance, a restriction also noted by 3C~\citep{zhang2024spoofdia}.
The bonafide attractor $\mathbf{a}_0$ is computed for each utterance from its own frames, so it adapts to the speaker and channel of that utterance.
With several bonafide speakers, one bonafide attractor may not be sufficient.
3C defines spoof-speaker diarization, spoof diarization with several speakers, and leaves it for future work.
Because the EDA comes from speaker diarization, it can be extended to one bonafide attractor per speaker; training such a model requires data with several speakers per utterance.

\paragraph{Non-speech.}
Following 3C, non-speech frames other than ConP have no training target and are not scored.
The behavior of the model on silence is therefore not defined.
This matters on a corpus without voice activity labels, such as LlamaPartialSpoof.
A deployment needs voice activity detection or training targets for non-speech.

\paragraph{Baselines.}
We did not retrain 3C or the attractor-token models under our protocol.
Their numbers are taken from the papers (App.~\ref{app:protocol}).
Our SSL controls (Table~\ref{tab:main}) remove part, but not all, of the front-end difference.

\paragraph{Run-to-run variation.}
The standard deviation of $\JER$ over runs is 0.3--2.4 points across configurations (0.95 for the default).
Differences below about 2 points between configurations are not resolved with three runs.

\paragraph{Future directions.}
A spoof diarization benchmark with several spoofing methods per utterance beyond PartialSpoof; spoof-speaker diarization with one bonafide attractor per speaker; training data with more speakers and labeled non-speech; estimation of $\Kmax$ from data; and an explicit non-speech class.

\section{Broader Impacts}
\label{app:impacts}

Spoof diarization can help forensic analysts find which parts of a recording were manipulated and group them by the method used.
Its output is not forensic proof.
E2E-SD misses a manipulation more often than it attributes it to a wrong method (App.~\ref{app:instances}).
Its performance also drops on data unlike PartialSpoof (Section~\ref{sec:transfer}).
A false alarm can harm the speaker of a genuine recording.
The outputs should therefore support an investigation and should not be the only evidence.
A published detector can also help an attacker test manipulations against it.
We release the code and the trained default model, and report its failure modes in App.~\ref{app:instances} and App.~\ref{app:error_cases}.

\section{Asset Licenses and Terms of Use}
\label{app:licenses}

Table~\ref{tab:licenses} lists the datasets, models and software that we use.
We use every asset for research only and under its license.
Our code is publicly available under the MIT license (App.~\ref{app:algorithms}); code adapted from other projects keeps its license (Table~\ref{tab:licenses}), and two EER functions taken from the ASVspoof 2019 t-DCF package keep its CC BY-NC-SA 4.0 license.
The trained default model is linked from the same address.

\begin{table}[H]
  \caption{Assets used in this work and their licenses.}
  \label{tab:licenses}
  \centering\small
  \newcommand{\lnk}[2]{{\footnotesize\def\UrlBreaks{\do\/\do\_}\href{#1}{\nolinkurl{#2}}}}
  \begin{tabular}{@{}>{\raggedright\arraybackslash}p{5.6cm}>{\raggedright\arraybackslash}p{3.7cm}>{\raggedright\arraybackslash}p{3.3cm}@{}}
    \toprule
    Asset & Use & License \\
    \midrule
    \multicolumn{3}{@{}l}{\textit{Data and pre-trained models}} \\
    ASVspoof 2019 LA~\tcite{todisco2019asvspoof,yamagishi2019asvspoof} & source of PartialSpoof; detection test & ODC-By \\
    PartialSpoof~\tcite{zhang2023partialspoof} & training, evaluation & CC BY 4.0 \\
    ASVspoof 2021 LA~\tcite{yamagishi2021asvspoof} & detection test & ODC-By \\
    ASVspoof 2021 DF~\tcite{yamagishi2021asvspoof} & detection test & ODbL 1.0 \\
    In-the-Wild~\tcite{mueller2022generalize} & detection test & CC BY-SA 4.0 \\
    LlamaPartialSpoof~\tcite{luong2025llamapartialspoof} & transfer test & CC BY 4.0 \\
    PartialEdit v1.1~\tcite{zhang2025partialedit} & transfer test & CC BY 4.0 \\
    WavLM-Large~\tcite{chen2022wavlm} & front-end & MIT \\
    wav2vec2-Large~\tcite{baevski2020wav2vec}, XLS-R 300M~\tcite{babu2022xlsr} & front-end controls & MIT (fairseq) \\
    \midrule
    \multicolumn{3}{@{}l}{\textit{Code in our repository (each part keeps its license)}} \\
    dscore~\tcite{dscore}\newline\lnk{https://github.com/nryant/dscore}{github.com/nryant/dscore} & scoring ($\JI$, $\JER$) & BSD 2-Clause \\
    PartialSpoof recipe\newline\lnk{https://github.com/nii-yamagishilab/PartialSpoof}{github.com/nii-yamagishilab/PartialSpoof} & gMLP back-end & BSD 3-Clause \\
    BUT EEND\newline\lnk{https://github.com/BUTSpeechFIT/EEND}{github.com/BUTSpeechFIT/EEND} & EDA & MIT \\
    DiariZen\newline\lnk{https://github.com/BUTSpeechFIT/DiariZen}{github.com/BUTSpeechFIT/DiariZen} & Conformer back-end (ablation) & MIT \\
    s3prl\newline\lnk{https://github.com/s3prl/s3prl}{github.com/s3prl/s3prl} & SSL model code (WavLM, wav2vec~2.0) & MIT \\
    ASVspoof 2019 t-DCF package\newline\lnk{https://www.asvspoof.org/resources/tDCF_python_v2.zip}{asvspoof.org/resources/tDCF_python_v2.zip} & two EER functions & CC BY-NC-SA 4.0 \\
    \bottomrule
  \end{tabular}
\end{table}

%%%%%%%%%%%%%%%%%%%%%%%%%%%%%%%%%%%%%%%%%%%%%%%%%%%%%%%%%%%%

\end{document}